\documentclass[10pt,twocolumn,letterpaper]{article}

\usepackage{wacv}      

\usepackage{graphicx}
\usepackage{amsmath}
\usepackage{amssymb}
\usepackage{booktabs}
\usepackage{url}
\usepackage{booktabs}
\usepackage{amsfonts}
\usepackage{nicefrac}
\usepackage{microtype}
\usepackage{xcolor}
\usepackage{times}
\usepackage{epsfig}
\usepackage{graphicx}
\usepackage{amsmath}
\usepackage{amssymb}
\usepackage{booktabs}
\usepackage{comment}
\usepackage{colortbl}
\usepackage{multirow}
\usepackage{xspace}
\usepackage{xr}

\usepackage{tabularx}
\usepackage{bbding}
\usepackage{cuted}
\usepackage{subcaption}
\usepackage[pagebackref,breaklinks,colorlinks]{hyperref}

\usepackage[capitalize]{cleveref}
\crefname{section}{Sec.}{Secs.}
\Crefname{section}{Section}{Sections}
\Crefname{table}{Table}{Tables}
\crefname{table}{Tab.}{Tabs.}

\def\wacvPaperID{1076} 
\def\confName{WACV}
\def\confYear{2025}

\begin{document}

\title{RefVSR++: 
Exploiting Reference Inputs \\
for Reference-based Video Super-resolution} 

\author{
Han Zou$^{1,2}$
~~~~Masanori Suganuma$^{1,2}$ 
~~~~Takayuki Okatani$^{1,2}$
\\
$^{1}$Graduate School of Information Sciences, Tohoku University
~~~~$^{2}$RIKEN Center for AIP \\
{\tt\small \{hzou, suganuma, okatani\}@vision.is.tohoku.ac.jp}
}

\maketitle

\begin{abstract}
{\color{black}
Smartphones with multi-camera systems, featuring cameras with varying field-of-views (FoVs), are increasingly common. This variation in FoVs results in content differences across videos, paving the way for an innovative approach to video super-resolution (VSR). This method enhances the VSR performance of lower resolution (LR) videos by leveraging higher resolution reference (Ref) videos. Previous works \cite{lee2022reference,kim2023efficient}, which operate on this principle, generally expand on traditional VSR models by combining LR and Ref inputs over time into a unified stream. However, we can expect that better results are obtained by independently aggregating these Ref image sequences temporally. Therefore, we introduce an improved method, RefVSR++, which performs the parallel aggregation of LR and Ref images in the temporal direction, aiming to optimize the use of the available data. RefVSR++ also incorporates improved mechanisms for aligning image features over time, crucial for effective VSR. Our experiments demonstrate that RefVSR++ outperforms previous works by over 1dB in PSNR, setting a new benchmark in the field.}

\end{abstract}

\section{Introduction}
\label{sec:intro}

Recent mobile phones are now equipped with multiple cameras, typically two or three, to enhance the overall photography experience. The equipped cameras have different focal lengths, allowing users to capture photos/videos with different fields of view (FoVs) simultaneously. For example, in a scenario where two cameras with different FoVs capture the same scene, the camera with a narrower FoV produces a higher-resolution image of the corresponding portion of the scene. Integrating it with the same scene image captured by the other camera with a larger FoV will allow us to increase its image resolution, at least in the region of overlap between the FoVs. 

This problem, known as reference-based image super-resolution (Ref-SR), has been studied recently in the community, resulting in the development of several methods \cite{lee2022reference,wang2021dual,huang2022task,zheng2018crossnet,xie2020feature,zhang2019image,zheng2017learning}. It is formalized as predicting a high-resolution SR image of a low-resolution (LR) input of a scene, aided by a reference (Ref) image of a similar scene. Specifically, we consider a scenario where the Ref image has a narrower FoV than the LR image. Existing Ref-SR methods first find the correspondences between the LR and Ref images, followed by warping the Ref image to align with the LR image, and finally fusing them to predict higher quality SR images.  

When taking a video, not a single photograph, of a scene with mobile phone cameras, we can increase the image resolution in a different way by using the images contained in the video. It is to aggregate the images in the temporal axis and obtain a rich visual feature of the captured scene, from which we super-resolve each image. Note that this is doable even with a single camera. This is called video SR (VSR), which has also been studied for a while in the community. 

{\color{black}
Now, with mobile phones equipped with multiple cameras, it is possible to integrate video SR with reference-based super-resolution (Ref-SR) to further enhance the quality of super-resolved images. Within a singular time step of video SR processing, the LR feature of an image can be enriched by transferring the detailed feature from the Ref image. Temporally, superior quality features from various Ref images can be disseminated via feature propagation. Consequently, this approach, known as reference-based video SR, can substantially improve the performance of video SR.

There are two key issues with reference-based video SR. The first pertains to how to make maximum use of all available data. While conventional video SR methods are adept at handling the LR image sequence, aiming to super-resolve these images, they may not be as effective with the Ref images. As shown in the Fig.~\ref{fig:ref_comp}, Ref images from different time steps provide superior details for different regions of a LR image. Features within the overlapped region appear refined and clear, while areas beyond the overlap turn coarse. Thus, given that an arbitrary number of Ref images at different time steps can provide critical information for super-resolving an LR image at a specific time step, it becomes essential to identify and integrate the most relevant Ref features for enhancing each LR feature in the sequence.

\begin{figure}[tb]
\centering
\includegraphics[width=0.95\columnwidth]{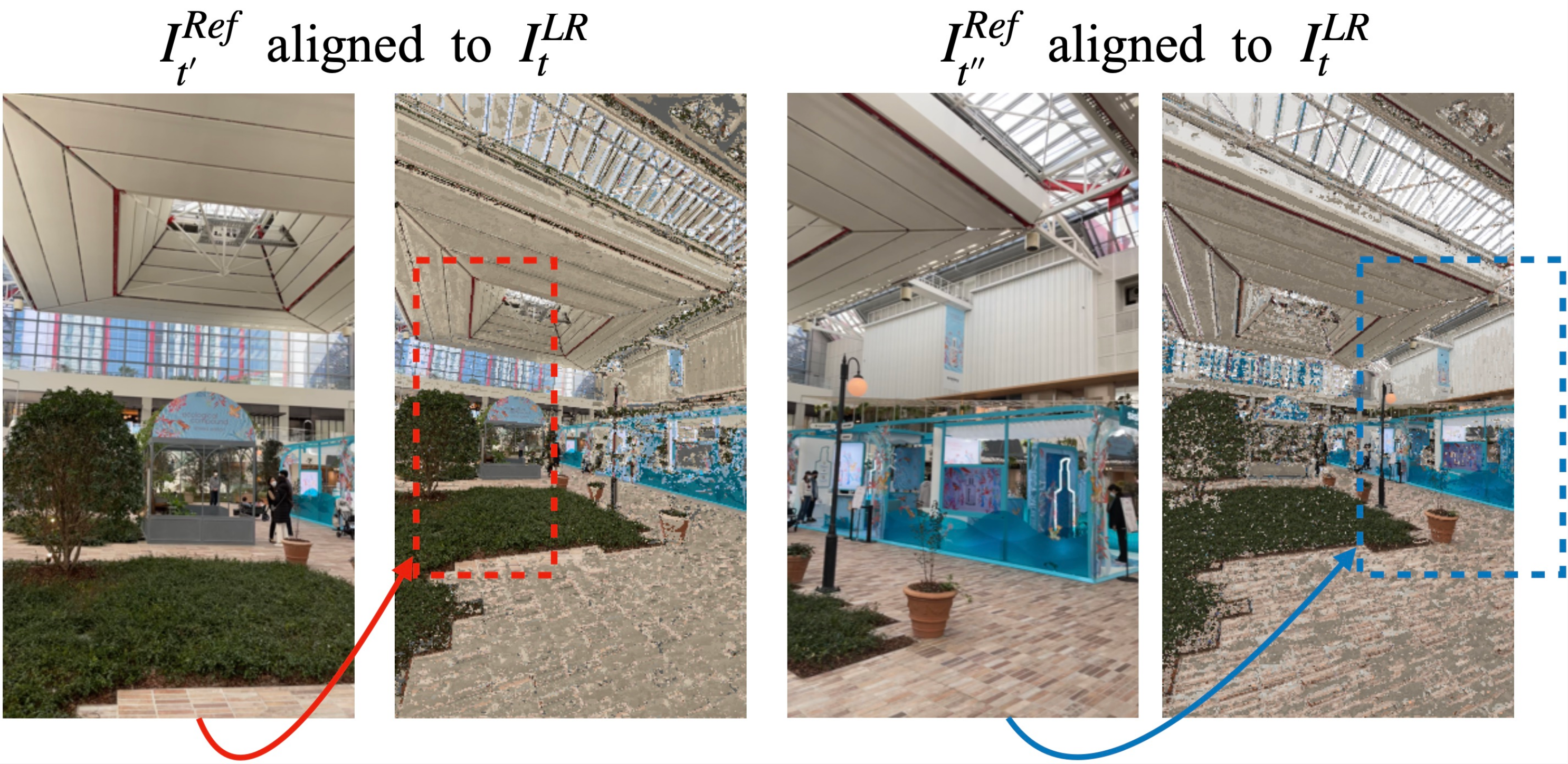}
\caption{{\color{black} Visualization of aligned Ref images from different time steps $I^{Ref}_{t'}$ and $I^{Ref}_{t''}$. By employing patch matching~\cite{barnes2009patchmatch}, matching patches are identified and aligned with the LR input $I^{LR}_{t}$.The areas marked by colored rectangles indicate regions of overlap. RGB images are utilized for better visualization.} }
\label{fig:ref_comp}
\vspace*{-2em}
\end{figure}

The second is how to align images accurately among the video sequence. Reference-based video SR requires aligning images over time and aligning the LR and Ref images at each time step. The former is a fundamental requirement of a well-established video SR approach, which propagates image features over time to generate high-quality SR images. However, the propagation is susceptible to the accumulation of errors, which causes a misalignment of images, leading to blurry SR outputs. We point out that existing methods for image alignment are error-prone. 

Based on these considerations, we propose a novel method for reference-based video SR dubbed RefVSR++. This method adopts the strategy of combining Video SR with Ref-SR methods and further expands on this concept to address the two primary challenges previously outlined. First, RefVSR++ is designed to maximize the use of both the Ref and LR image sequences. Specifically, RefVSR++ utilizes two streams to propagate different features. One stream propagates the features of Ref images, referred to as the Ref stream, and the other propagates aggregated Ref and LR image features, referred to as the SR stream. Such a design could better aggregate and propagate various Ref features among the video sequence. Second, RefVSR++ addresses the issue of image alignment by utilizing a more accurate and robust method compared to the previous works. 

These component methods enable effective feature aggregation across all input images over an extended period of time, leading to superior SR outputs. The resultant method, RefVSR++, has established the new state-of-the-art on the RealMCVSR dataset~\cite{lee2022reference}. }

\section{Related Work}
\subsection{Video Super-Resolution }

The goal of video super-resolution (VSR) is to recover a high-resolution (HR) video from a low-resolution (LR) input video. VSR has been studied for a long time. Recent methods are classified into two categories: methods based on a sliding window and those based on recurrent computation. The sliding window is widely used in CNN-based VSR methods~\cite{shi2016real, zhu2019residual,li2020mucan,tian2020tdan, wang2019edvr,jo2018deep}, which receive several consecutive frames as input, traverse them with a sliding window, and then predict an SR image of their center frame. These methods must process each frame multiple times to handle a long video. Thus, they tend to suffer from a high computational cost, and they can deal with a limited number of input frames, which makes it hard to deal with long-term dependencies. Methods based on recurrent computation~\cite{huang2017video,isobe2020video,haris2019recurrent,chan2021basicvsr,chan2022basicvsr++,sajjadi2018frame} utilize reconstructed high-quality images at previous time steps or their features to generate high-quality images at the current time step. Huang et al.~\cite{huang2017video} propose recurrent bi-directional networks to better utilize temporal information. Chan et al.\cite{chan2022basicvsr++} propose using high-order grid connections and flow-guided alignment for the recurrent computation. 

\subsection{Reference-based Super-Resolution}

Reference-based super-resolution (RefSR) uses additional reference image(s) to super-resolve an input low-resolution image. Previous studies have shown the effectiveness of transferring information from a high-resolution reference image to generate SR images. A critical problem is accurately aligning the Ref image with the LR image, which is important for fusing their image features in a subsequent step to generate high-quality SR images. Zheng et al.~\cite{zheng2018crossnet} estimate optical flows between them for their alignment. Zhang et al.~\cite{zhang2019image} propose using patch matching~\cite{barnes2009patchmatch}, and Yang et al.~\cite{yang2020learning} improve it by adopting attention mechanisms for feature fusion. Wang et al.~\cite{wang2021dual} propose an aligned attention method for better feature fusion, which preserves high-frequency features via spatial alignment operations well. Huang et al.~\cite{huang2022task} decouple ref-based SR task into two sub-tasks: single image SR task and texture transfer task, and train them independently. It reduces misuse and underuse of the Ref feature, which often happens in Ref feature transfer. Lee et al.~\cite{lee2022reference} propose RefVSR, which integrates RefSR with VSR; we will discuss their method in detail below. Kim et al.~\cite{kim2023efficient} propose ERVSR, an enhancement of RefVSR that utilizes only a single reference image, thereby improving resource efficiency.

\section{Improved Use of References for VSR}

\subsection{Problem Formulation and Notation}

Let $I^{LR}_t\in\mathbb{R}^{H\times W\times C}$ and $I^{Ref}_t\in\mathbb{R}^{H\times W\times C}$, where $t=1,\ldots,T$, represent the input sequences of low-resolution (LR) and reference (Ref) images, respectively. Here, $H$, $W$, and $C$ denote the height, width, and number of channels in the images, respectively. Note that $I^{LR}_t$ captures a scene with a large field of view (FoV), while $I^{Ref}_t$ represents an image of the same scene but with a narrower FoV. This makes $I^{Ref}_t$ a higher-resolution image of a specific portion of $I^{LR}_t$ (refer to Fig.~\ref{fig:8k_qua}). The objective is to produce a super-resolution (SR) version $I^{SR}_t\in\mathbb{R}^{sH\times sW\times C}$ of $I^{LR}_t$, using a predetermined upscaling factor $s$.

\begin{figure}[t]
\centering
\includegraphics[width=0.95\columnwidth]{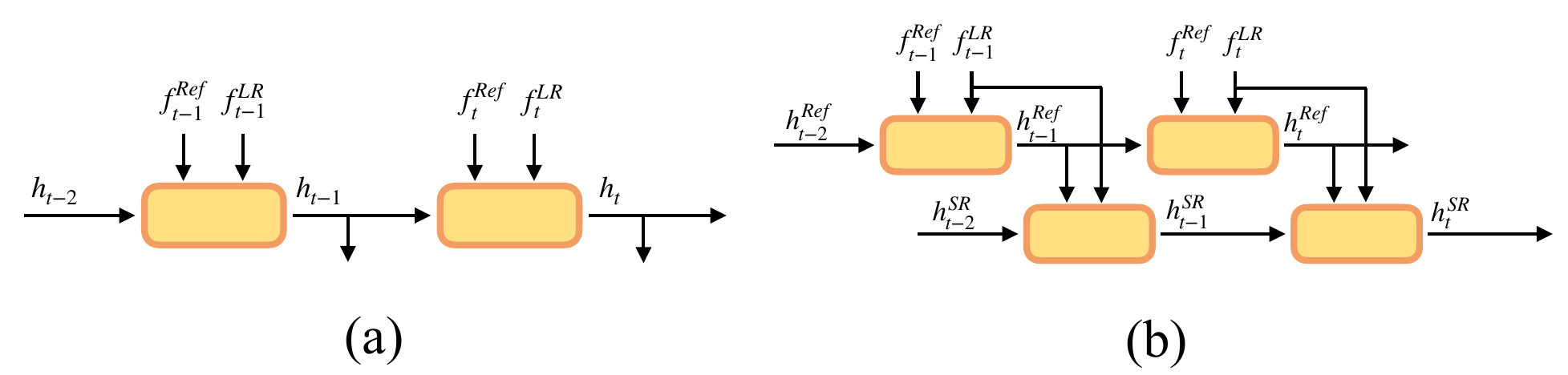}
\caption{Comparison of (a) RefVSR and (b) RefVSR++(ours). While RefVSR propagates only a single feature, ours propagates two features to aggregate all available information over time. For simplicity, only the forward propagation is shown.} 
\label{fig:comp}
\vspace*{-1em}
\end{figure}

To solve the above problem, RefVSR~\cite{lee2022reference} integrates reference-based SR and video SR in the following way. First, it utilizes the Ref image $I^{Ref}_t$ at each time step $t$ by warping its feature (map) $f^{Ref}_t$ and fusing it with a scene image feature $h_t$ computed at $t$ as follows.  \cite{yang2020learning,zhang2019image,zheng2018crossnet,wang2021dual}. To compute the features $h_t$, RefVSR aggregates the feature $f^{LR}_t$ of $I^{LR}_t$ $(t=1,\ldots,T)$ over time by recurrently propagating $h_t$ with $t=1,\ldots,T$, following the established methods of video SR \cite{huang2015bidirectional,huang2017video,haris2019recurrent,chan2021basicvsr,chan2022basicvsr++}.
To be specific, $h_{t-1}$ from the previous time step is aligned with the current feature $f^{LR}_{t}$ to compensate for the motion from $t-1$ to $t$, and then fused with $f^{LR}_{t}$ to update it to $h_{t}$. The updated feature $h_t$ is used to generate a SR image $I^{SR}_t$; see Fig.~\ref{fig:comp}(a). 

\begin{figure}[t]
\centering
\includegraphics[width=0.95\columnwidth]{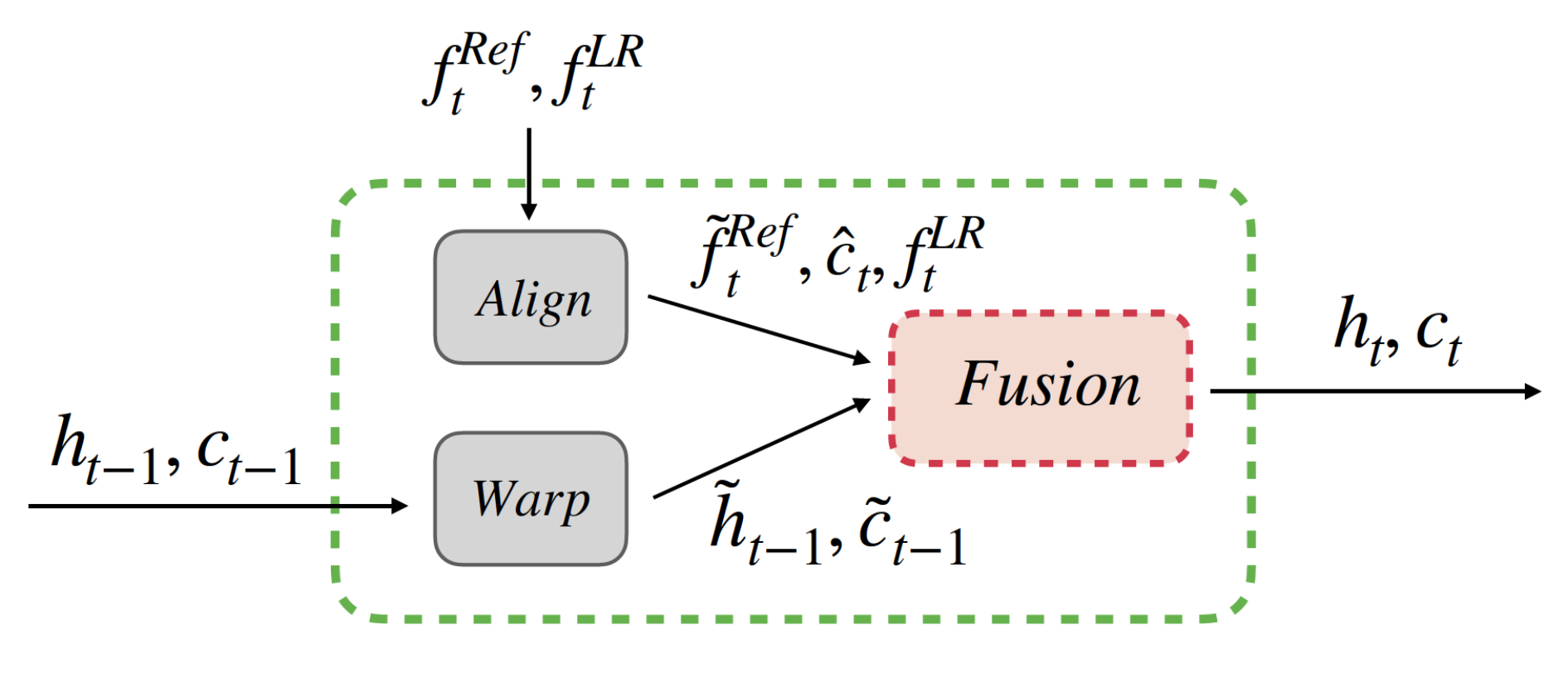}
\caption{Design of the recurrent cell of RefVSR.}
\label{fig:RefVSR_cell}
\vspace*{-1em}
\end{figure}

The computation in each recurrent cell is as follows; see Fig.~\ref{fig:RefVSR_cell}. It retains $c_t$, a confidence map for $h_t$, and propagates it together with $h_t$. Given an input pair $h_{t-1}$ and $c_{t-1}$ from $t-1$ as input, the recurrent cell warp them into $\tilde{h}_{t-1}$ and $\tilde{c}_{t-1}$ to align  with $f^{LR}_t$. It also aligns the Ref feature $f^{Ref}_t$ with $f^{LR}_t$, yielding the aligned $\tilde{f}^{Ref}_t$ with a confidence map $\hat{c}_t$ for the alignment. Next, $\tilde{h}_{t-1}$, $\tilde{f}^{Ref}_t$, and $f^{LR}_t$ are fused to produce $h_t$ (and its fusion confidence $c_t$). Here, a confidence-guided fusion is performed using confidence map $\hat{c}_t$ to select well-matched features from $\tilde{f}^{Ref}_t$ and using $\tilde{c}_{t-1}$ to select well-matched features from $\tilde{h}_{t-1}$.

There exist two issues with this approach. Firstly, 
 {\bf it does not 
utilize the Ref inputs properly.
At each time step $t$, two observations, $I^{Ref}_t$ and $I^{LR}_t$, are fused into $h_t$ and carried forward over time;
see Fig.~\ref{fig:comp}(a). The information from $I^{Ref}_t$ and $I^{LR}_t$ is first merged and then progressively accumulated.}
However, $I^{Ref}_t$ and $I^{LR}_t$ contain different information due to their FoVs.  The aligned Ref feature $\tilde{f}^{Ref}_t$  outside the overlapped FoV is error-prone, leading to possible degradation outside the overlapped FoV. A more proper approach is to  propagate and accumulate
the LR and Ref features independently over time, allowing for the extraction of high-frequency propagated Ref features from adjacent error-prone features.

Secondly, {\bf RefVSR propagates a confidence map $c_t$ along with $h_t$ over time using an heuristic approach, accumulating errors in the propagation.} The goal is to select well-matched features for feature fusion. However, the method for updating the confidence map is heuristic; it involves taking the greater of the warped propagated confidence map $\tilde{c}_{t-1}$ and the confidence map $\hat{c}_t$ from the alignment of $f^{Ref}_t$ and $f^{LR}_t$. The warping is based on estimates of optical flow between images, but these estimates can be inaccurate. This inaccuracy may lead to cumulative errors, potentially compromising the accuracy of Ref feature fusion.

\subsection{Proposed Approach}

Considering the above points, we propose adding an extra stream that aggregates information from the Ref images over time, as depicted in Fig.~\ref{fig:comp}(b). This stream propagates the feature $h^{Ref}_t$ from $I^{Ref}_t$ through time independently of the main stream, aiming to overcome the first limitation of RefVSR. The main stream fuses the Ref feature $f^{Ref}_t$ and the LR feature $f^{LR}_t$ and propagates the fused feature in a manner that enhances the original RefVSR approach. We refer to this fused feature as $h^{SR}_t$, as it is utilized to generate the SR output $I^{SR}_t$.

To address the second limitation of RefVSR, we introduce a more effective feature fusion technique in both streams, eliminating the need for propagating a confidence map. Further details on this will be provided later. It's important to note that, consistent with RefVSR and other modern VSR methods, feature propagation in each stream is performed bidirectionally.

\section{Details of the Proposed Method}

\subsection{Overview}

\begin{figure}[bt]
\centering
\includegraphics[width=0.95\columnwidth]{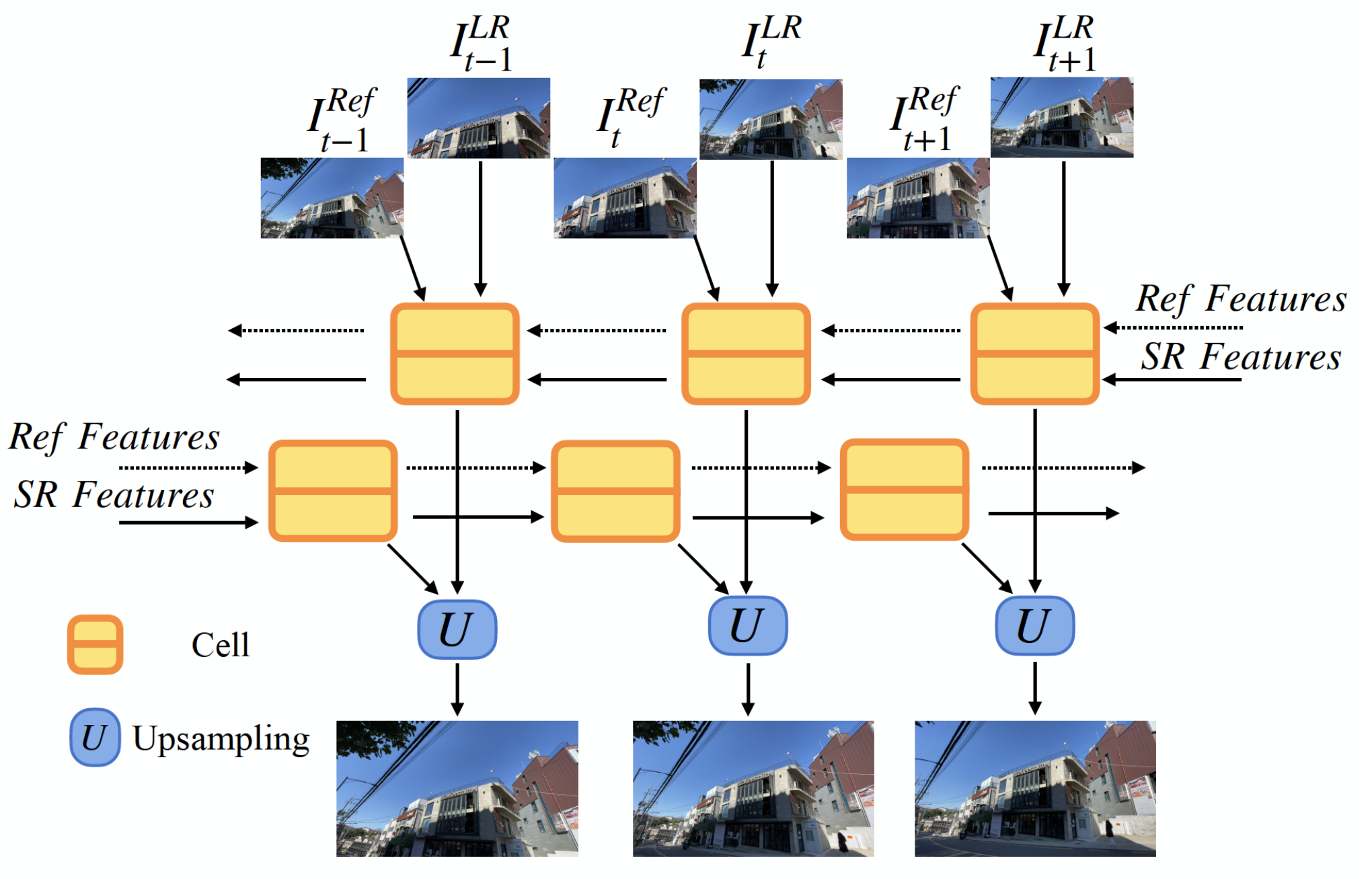}
\caption{Overview of the proposed method. $I^{LR}_t$ and $I^{Ref}_t$ are inputted into  the 
recurrent cells in forward and backward directions. The arrows to the cells in the forward stream are omitted for brevity.}
\label{fig:framework}
\end{figure}

Figure \ref{fig:framework} shows the overview of the proposed network. As also shown in Figs.~\ref{fig:comp}(b), it has two recurrent cells that propagate and update $h^{Ref}$ and $h^{SR}$ in each of the forward and backward streams and a single module that fuses the two $h^{SR}$'s from these streams and generates $I^{SR}$. The latter is merely an upsampling module consisting of ResBlocks~\cite{he2016deep} and pixel shuffle~\cite{shi2016real} as in RefVSR. 

We use forward propagation for an explanation below and omit the backward propagation since we do the same for it in the opposite direction. In each recurrent cell, we first extract the feature $f^{LR}_{t}$ and $f^{Ref}_{t}$ from $I^{LR}_{t}$ and $I^{Ref}_{t}$ using two encoders, respectively. 

\subsection{Updating Ref Features}\label{sec:ref_cell}

\begin{figure}[tb]
\centering
\includegraphics[width=0.95\columnwidth]{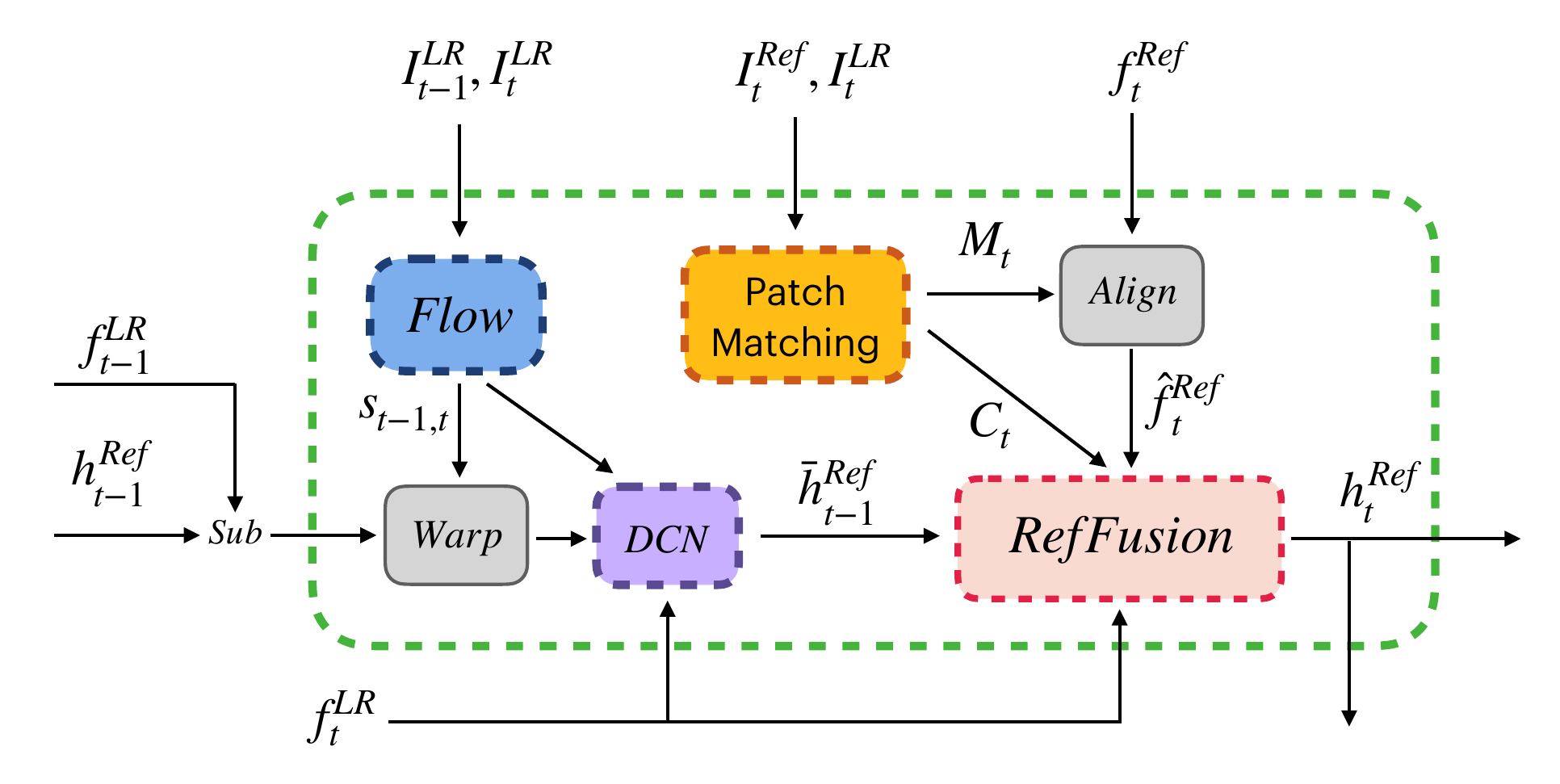}
\caption{The recurrent cell for the Ref feature stream}
\label{fig:cell_1}
\end{figure}

Figure \ref{fig:cell_1} shows the recurrent cell for the Ref feature stream. The cell updates $h^{Ref}_{t}$ as follows:
\begin{equation}
\label{eqn:refcell}
    h^{Ref}_{t} = F^{Ref}(h^{Ref}_{t-1}; f^{Ref}_{t}, f^{LR}_{t}, s_{t-1,t}),
\end{equation}
where $s_{t-1,t}$ is the optical flow from $I^{LR}_{t-1}$ to $I^{LR}_t$ estimated by a flow estimator~\cite{ranjan2017optical}. 

Given the input feature $h^{Ref}_{t-1}$ received from the previous step, the process begins by warping it to align the current LR image $I^{LR}_t$. As per RefVSR, a flow estimator is employed to calculate the optical flow $s_{t,t-1}$ between $I^{LR}_{t-1}$ and $I^{LR}_t$,  which is then used for the alignment. 

Now, let $\breve{h}^{Ref}_{t-1}$ be the resulting warped feature, which is omitted in Fig.~\ref{fig:cell_1}. As potential misalignment due to errors in the estimated flow $s_{t-1,t}$ can make $\breve{h}^{Ref}_{t-1}$ inaccurate, we need more fine-grained alignment. We then employ DCN as follows:
\begin{equation}
    \bar{h}^{Ref}_{t-1} = \mathcal{D}^{Ref}(\breve{h}^{Ref}_{t-1}, o^{Ref} ,m^{Ref}),
    \label{eqn:first}
\end{equation}
where $o^{Ref}$ is offset for the estimated optical flow and $m^{Ref}$ is a modulation mask, which are computed as 
\begin{subequations}
\begin{align}
    o^{Ref} &= C^{Ref}_{o}([f^{LR}_{t}, \breve{h}^{Ref}_{t-1}]) + s_{t-1,t } , \\
    m^{Ref} &= \sigma (C^{Ref}_{m}([f^{LR}_{t}, \breve{h}^{Ref}_{t-1}])),
\end{align}
\end{subequations}
where $C^{Ref}_{o}$ and $C^{Ref}_{m}$ are convolutional layers and $\sigma$ is a sigmoid function.
To adjust for the field of view (FoV) differences, we align $f^{Ref}_t$ with $f^{LR}_t$ adaptively, based on the content of the images.  We use {\em patch matching}, a method supported by existing studies\cite{lee2022reference,wang2021dual,huang2022task}, which has demonstrated superiority in warping Ref features compared to warping based on optical flow. Specifically, it computes an index map $M_t$ with its confidence map $C_t$ for alignment as
\begin{equation}
    \{M_{t}, C_{t}\} = Match(I^{LR}_{t}, I^{Ref}_{t}).
\end{equation}
Specifically, we first embed $I^{Ref}_t$ and $I^{LR}_t$ into feature maps and extract $3\times3$ patches with a stride $1$ using a shared encoder. Then we calculate the cosine distance $S_{i,j}$ between pairs of the LR feature patch $i$ and the Ref feature patch $j$. The matching index and its confidence for patch $i$ is given by  $M_{t,i} = \arg \max_{j} S_{i,j}$ and $C_{i} = \max_{j} S_{i,j}$, respectively. Finally, we warp $f^{Ref}_{t}$ to $\hat f^{Ref}_{t}$ using the index map $M_t$.

We then update the propagated Ref feature $\bar{h}^{Ref}_{t-1}$ with the warped Ref feature $\hat f^{Ref}_t$. There are two potential issues with fusing such Ref features from different time steps. First, they may have different sharp textures at the same position due to the change of depth of field or illumination. Second, they contain different reliable regions. Thus, their direct fusion may result in a mismatch or blurred feature. To obtain a sharp fused Ref feature and preserve information from the current and previous Ref frames, we selectively transfer textures from the two Ref features $\bar h^{Ref}_{t - 1}$ and $\hat f^{Ref}_t$ to the LR feature.

Specifically, we fuse each of $\bar{h}^{Ref}_{t-1}$ and $\hat f^{Ref}_t$ with the current frame's LR feature $f^{LR}_t$ separately, aiming to preserve textures from the Ref features from different sources. 
\begin{subequations}
\begin{align}
    f_{t}^{a}  &= Conv([f^{LR}_t, \bar{h}^{Ref}_{t-1}],\\
    f_{t}^{b}  &= Conv(C_{t}) \cdot Conv(f^{LR}_t, \hat f^{Ref}_t).
\end{align}
\end{subequations}
where $Conv$ is a convolutional layer. We adopt an adaptive fusion guided by the confidence map $C_t$ to fuse $\hat f^{Ref}_t$. We then get the Ref feature by fusing the two Ref features obtained above as 
\begin{equation}
    h^{Ref}_{t}= \mathcal{R}(f_{t}^{a}, f_{t}^{b}, f^{LR}_t).
\end{equation}
where $\mathcal{R}$ is ResBlocks. 
We use $ h^{Ref}_{t}$ to refine the propagated SR feature in the SR feature cell. Finally, we obtain the Ref feature as in (\ref{eqn:refcell}), which will be propagated to the next time step.

\subsection{Ref Residual Propagation}
\label{sec:residual}

The Ref cell transfers the learned Ref feature to the subsequent time step. To maintain high-frequency details throughout the propagation, we convey the Ref residual rather than the fused feature. Given that the LR feature $f^{LR}_t$ contains coarser features in contrast to those derived from the Ref frame, the residual is computed as follows:
\begin{align}
    h^{Ref}_t \gets h^{Ref}_t - f^{LR}_t
\end{align}
Note also that the `low-frequency' component, missing in the residual, is supplemented by the current frame's $f^{LR}_t$ and is 
fused at a later stage. Therefore, opting for the residual Ref features does not affect other designs. This  method is adopted in the experiments unless otherwise noted; see  Sec.~\ref{sec:ablation} for its experimental verification.

\subsection{Updating SR Features } 
\begin{figure}[tb]
\centering
\includegraphics[width=0.95\columnwidth]{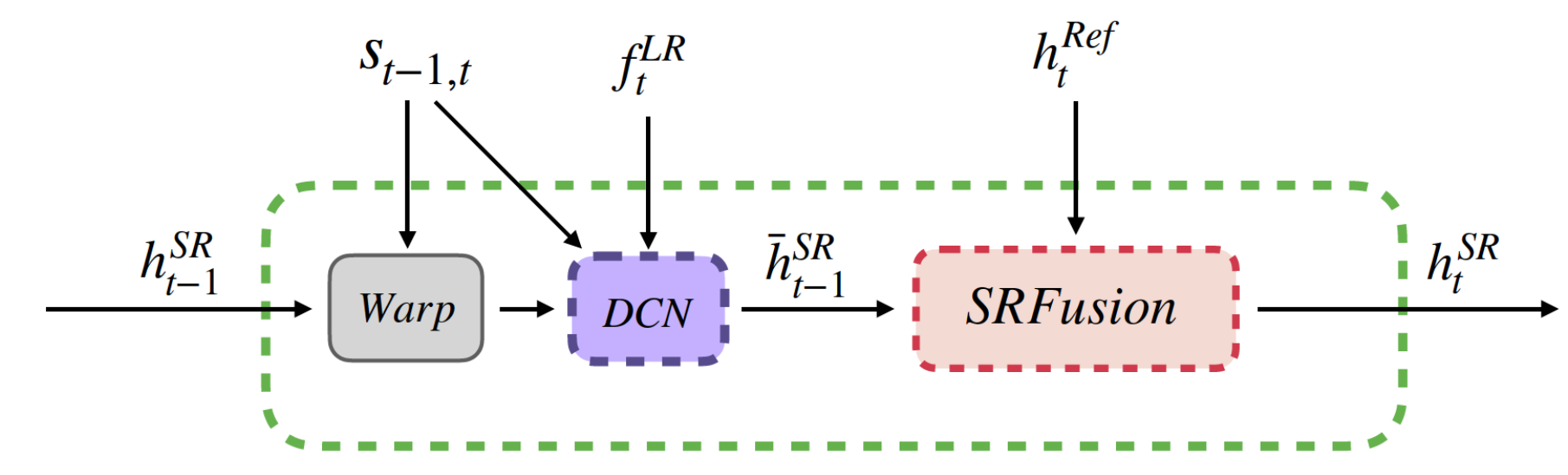}
\caption{The recurrent cell for the SR feature stream.}
\label{fig:cell_2}
\end{figure}
Figure \ref{fig:cell_2} shows the recurrent cell for the SR feature stream. The cell updates the propagated SR feature using the Ref feature $h^{Ref}_t$ computed above and the current LR feature $f^{LR}_t$ as
\begin{equation}
    h^{SR}_{t} = F^{SR}(h^{SR}_{t-1}, h^{Ref}_t, f^{LR}_{t}, s_{t-1,t}).
\end{equation}

First, we align the propagated LR feature $h^{SR}_{t-1}$ to the LR feature of the current time step $t$. Similarly to the Ref feature cell, we first warp it with the estimated flow $s_{t-1,t}$ and then apply deformable convolution~\cite{dai2017deformable} to cope with possible misalignment. Letting $\breve{h}^{SR}_{t-1}$ be the warped feature by $s_{t,t-1}$, we apply DCN as
\begin{equation}
\bar{h}^{SR}_{t-1} = \mathcal{D}^{SR}(\breve{h}^{SR}_{t-1}, o^{SR}, m^{SR}), 
\end{equation}
with offset $o^{SR}$ of flows and a modulation mask $m^{SR}$, which are given by
\begin{subequations}
\begin{align}
    o^{SR} &= C^{SR}_{o}(h^{Ref}_{t}, \breve{h}^{SR}_{t-1})+ s_{t-1,t},\\   
    m^{SR} &= \sigma (C^{SR}_{m}(h^{Ref}_{t}, \breve{h}^{SR}_{t-1})),
\end{align}
\end{subequations}
where $C^{SR}_{o}$ and $C^{SR}_{m}$ are convolutional layers and $\sigma$ is a sigmoid function. 

Next, we fuse the aligned feature $\breve{h}^{SR}_{t-1}$ with the updated Ref feature $h^{Ref}_{t}$ from the other cell using several ResBlocks as 
\begin{equation}\label{eqn:second}
    h^{SR}_{t} = \mathcal{R}(h^{Ref}_{t}, \breve{h}^{SR}_{t-1} ) +\breve{h}^{SR}_{t-1}.
\end{equation}
The resulting $h^{SR}_{t}$ will be used to generate SR output at the current time step $t$ and also propagated to the next time step.

\subsection{Training}
\label{sec:training}

We use the RealMCVSR dataset \cite{lee2022reference} in our experiments. It consists of ultra-wide, wide-angle, and telephoto videos from multiple scenes. The three videos have the same size but different FoVs. 
Note that the wide and telephoto videos have twice ($2\times$) and four times ($4\times$) the magnification of the ultra-wide video, respectively. 
Following the paper\cite{lee2022reference}, we consider the $4\times$ super-resolution of an ultra-wide video using a wide-angle video as a Ref video. This yields an 8K video of the scene. 

We use the two-stage training strategy used in previous studies~\cite{wang2021dual,lee2022reference}. Specifically, we first train the proposed model in a supervised fashion. In this stage, $4\times$ down-sampled ultra-wide and wide-angle videos are treated as LR and Ref inputs, respectively, and the original ultra-wide videos are treated as ground-truth. Subsequently, the pre-trained model is fine-tuned to adapt it to the original-sized videos. In this stage, the original ultra-wide and wide-angle videos serve as LR and Ref inputs, respectively. Due to the absence of ground-truth 8K ultra-wide videos, we use an approximate loss by using the original wide-angle and telephoto videos as pseudo ground-truths. 

The details of the first stage are as follows. 
We train the model to use the $4\times$ downsampled versions of $I^{UW}_t$ and $I^{Wide}_t$ as $I^{LR}_t$ and $I^{Ref}_t$, respectively, to predict as close $I^{SR}_t$ to the original high-resolution $I^{UW}_t$ as possible. 
We use the weighted sum of two loss functions, the reconstruction loss and the reference fidelity loss:
\begin{equation}
     \ell_{\mathrm{1st}} = \ell_{rec} + \beta \ell_{fid},
     \label{eqn:loss1st}
\end{equation}
where $\beta$ is a weighting constant. 

For $\ell_{rec}$, we use the same loss as RefVSR, which evaluates the low-frequency and high-frequency components of $I^{SR}$ in two terms, respectively, as
\begin{equation}
     \ell_{rec} = ||I^{SR}_{t,blur}-I^{UW}_{t,blur}|| + \alpha \sum_{i}\delta_i(I^{SR}_{t}, I^{UW}_t),
\end{equation}
where $I_{t,blur}$ indicates $I_t$ is filtered by $3\times3$ Gaussian kernels with $\sigma=0.5$. The second term is called the contextual loss; $\delta_i(X, Y)=min_j\mathbb{D}(x_i,y_j)$ measures the distance between pixel $x_i$ and the most similar pixel $y_j$ at a perceptual distance $\mathbb{D}$. 

For $\ell_{fid}$, we employ the fidelity loss proposed in \cite{wang2021dual} that uses only a single Ref frame as
\begin{equation}
     \ell_{fid}(I^{SR}_t,I^{Wide}_t) = \frac{\sum_{i} \delta_i (I^{SR}_{t},I^{Wide}_{t})\cdot c_i}{\sum_i c_i},
\end{equation} 
where $c_i$ is matching confidence. 

The details of the second training stage are as follows. To overcome the performance drop due to the fact that the model is trained on the down-sampled input, we follow the fine-tuning strategy proposed in \cite{wang2021dual}. Specifically, we use the original $I^{UW}_t$ and $I^{Wide}_t$ as the LR and Ref inputs, respectively. As there is no ground truth in this case, we $4\times$ down-sample the predicted $I^{SR}_t$ and measure its difference from $I^{UW}_t$. In addition, we also use the fidelity loss between $I^{SR}_t$ and the corresponding image $I^{Tele}_t$ of the telephoto video. Therefore, the loss is given as follows:
\begin{equation}
    \ell_{\mathrm{2nd}} = ||I^{SR}_{t\downarrow,blur}-I^{UW}_{t,blur}|| + \gamma \ell_{fid}(I^{SR}_t,I^{Tele}_t),
    \label{eqn:loss2nd}
\end{equation}
where $\gamma$ is a weighting constant.

\section{Experimental Results}

We evaluate our method using the RealMCVSR dataset \cite{lee2022reference}. It contains 161 triplets (i.e., ultra-wide, wide-angle, and telephoto, as explained in Sec.~\ref{sec:training}) of HD-resolution ($1080\times 1920$) videos captured by iPhone 12 Pro Max equipped with triple cameras. 
The dataset is divided into training, validation, and testing sets, with 137, 8, and 16 triplets, respectively. We follow \cite{lee2022reference} for the experimental settings and thus omit the details here; see them in the supplementary. Following previous studies~\cite{wang2021dual, lee2022reference}, we set the loss weights $\alpha$, $\beta$, and $\gamma$ to 0.01, 0.05, and 0.1, respectively. 

\subsection{Quantitative Evaluation}
\setlength{\tabcolsep}{3pt}
\begin{table}[bt]
\centering \footnotesize 
\begin{tabular}{lcccc} 
\hline
Method &  Type & PSNR(dB) & SSIM & Params \\\cline{2-3}
\hline
Bicubic                           & -  & 26.65 & 0.800 & - \\
RCAN-$pix$~\cite{zhang2018rcan}    & SR & 31.07 & 0.915 & 15.89M\\
IconVSR-$pix$~\cite{chan2021basicvsr} & VSR &33.80 & 0.951 & 7.26M\\ 
BasicVSR++-$pix$~\cite{chan2022basicvsr++} & VSR  &32.80 &0.941 &7.32M \\
TTSR~\cite{yang2020learning}       &R-SR & 30.31 & 0.905 &6.7M \\
TTSR-$pix$~\cite{yang2020learning} &R-SR & 30.83 & 0.911 &6.7M\\
DCSR~\cite{wang2021dual}           &R-SR & 30.63 & 0.895 &5.42M\\
DCSR-$pix$~\cite{wang2021dual}     &R-SR & 32.43 & 0.933 &5.42M\\
ERVSR-$pix$~\cite{kim2023efficient}     &R-VSR & 34.44 & 0.957 & 9.16M \\
RefVSR~\cite{lee2022reference}     &R-VSR & 31.73 & 0.916 &4.78M\\
RefVSR-$pix$~\cite{lee2022reference} &R-VSR& 34.86 & 0.959 &4.78M\\
RefVSR++-$small$               &R-VSR&  32.02  & 0.918  &3.24M \\
RefVSR++-$small$-$pix$                &R-VSR&  35.23  & 0.959   &3.24M \\
RefVSR++                        &R-VSR& 32.26 &  0.925 & 7.22M\\
RefVSR++-$pix$                   &R-VSR& \textbf{35.90}  & \textbf{0.965}  &7.22M \\
\hline
\end{tabular}
\caption{Quantitative evaluation of methods on the RealMCVSR dataset. Method column: `-$pix$' indicates the method is trained with a pixel-based loss; see the texts for details. Type column: SR means single-image SR and VSR means video SR. `R-' indicates it is reference-based. PSNR and SSIM~\cite{wang2004image} values for all methods except our own were obtained from \cite{lee2022reference} and \cite{kim2023efficient}. The `Params' column lists the number of parameters in millions.}
 \vspace{-1em}
\label{table:quan}
\end{table}

\begin{table*}[t]
\centering 
\begin{tabular}{lcccccc}
\hline
Model &  0-50\% & 50\%-60\%& 50\%-70\%& 50\%-80\%& 50\%-90\% & 50\%-100\%   \\
\hline
Bicubic                                 & 25.38/0.757 & 26.30/0.785 & 26.42/0.789 & 26.71/0.798 & 26.99/0.801 & 27.29/0.815 \\
RCAN~\cite{zhang2018rcan}         & 29.77/0.895 & 30.69/0.908 & 30.86/0.910 & 31.17/0/914 & 31/50/0.918 & 31.80/0.921 \\
IconVSR~\cite{chan2021basicvsr}  & 32.79/0.946 & 33.43/0.949 & 33.60/0.950 & 33.89/0.951 & 34.19/0.953 & 34.40/0.953 \\
DCSR~\cite{wang2021dual}          & 34.90/0.963 & 31.96/0.927 & 31.61/0.921 & 31.58/0.919 & 31.81/0.921 & 31.93/0.923({\color{blue}$-8.5\%$/$-4.2\%$})  \\
RefVSR~\cite{lee2022reference}    & 36.14/0.971 & 34.66/0.959 & 34.40/0.956 & 34.34/0.955 & 34.52/0.955 & 34.63/0.955({\color{blue}$-4.2\%$/$-1.6\%$})  \\ 
RefVSR++-$small$ & 36.18/0.971 & 35.16/0.962 & 34.95/0.960 & 34.88/0.959 & 35.02/0.959 & 35.08/0.958({\color{blue}$-3.0\%$/$-1.3\%$}) \\
RefVSR++  & 37.17/0.976 & 35.99/0.967 & 35.70/0.965 & 35.53/0.963 & 35.63/0.963 & 35.66/0.962({\color{blue}$-4.1\%$/$-1.4\%$}) \\ 

\hline
\end{tabular}
\caption{Quantitative results (PSNR (dB)/SSIM) measured with centered image regions with different sizes. {\color{black}The drop ratio of Ref-based methods within the 50\%-100\% FoV range is highlighted in blue text. See texts for details.}}
\vspace{-3mm}
\label{table:fov}
\end{table*}

\setlength{\tabcolsep}{3pt}
\begin{table}[bt]
\centering \footnotesize 
\begin{tabular}{lcccc} 
\hline
Method &  Type & PSNR(dB) & SSIM & Params  \\\cline{2-3}
\hline
ERVSR-$pix$~\cite{kim2023efficient}     &R-VSR & 27.12 & 0.886   & 9.16M \\
RefVSR-$pix$~\cite{lee2022reference}    &R-VSR & 27.91 & 0.869   &4.78M\\
RefVSR++-$pix$                          &R-VSR & \textbf{29.73} & \textbf{0.892}   &7.22M \\
\hline
\end{tabular}
\caption{Quantitative evaluation of methods on the selected video clips from Inter4K~\cite{stergiou2021adapool} dataset.}
\label{table:inter4k}
\vspace{-2mm}
\end{table}

\setlength{\tabcolsep}{2pt}
\begin{table}[bt]
\centering 
\fontsize{8pt}{3mm}\selectfont
\begin{tabular}{ccccccccc}
\hline
 No.  & w/ Ref & Conf. & SR stream & Ref stream  & Res & PSNR & Params & Time(s)\\
\hline
1  &&&&& & 34.02 &4.43M& 0.053\\ 
2 & \CheckmarkBold && & &&   35.19& 5.48M& 0.106 \\
3 & \CheckmarkBold&\CheckmarkBold && &&   35.15& 5.49M & 0.108\\
4 &  \CheckmarkBold&&\CheckmarkBold& && 35.72& 6.24M & 0.116 \\
5 & \CheckmarkBold& & &  \CheckmarkBold && 35.45& 6.35M& 0.112\\
6 & \CheckmarkBold& & & \CheckmarkBold & \CheckmarkBold & 35.62 &6.35M& 0.114 \\
7 & \CheckmarkBold& &\CheckmarkBold&\CheckmarkBold&  \CheckmarkBold & 35.90 & 7.22M& 0.126\\
\hline

\end{tabular}
\caption{Results of the ablation. See the text for details. }
\label{table:ablation_dual}
\vspace*{-1em}
\end{table}

\begin{table*}[tbh]
\centering 
\vspace{-2mm}
\begin{tabular}{lccccc}
\hline
Method & Time(s/frame)  & MACs(G/frame) & Params & Memory(GB)\\ 
\hline

RefVSR~\cite{lee2022reference}  & 1.036   &  7681.84 & 4.28M & 19.238 \\ 
ERVSR~\cite{kim2023efficient} & 0.888  & 1049.78 & 9.16M &5.378\\ 
RefVSR++-$small$ & 0.912  & 427.01  & 3.24M & 18.308 \\ 
RefVSR++         & 0.126  & 1000.48 & 7.22M & 19.408\\ 
\hline
\end{tabular}
\caption{Comparison of computational cost. }
\label{table:cost}
\vspace{-2mm}

\end{table*}

We first quantitatively evaluate the proposed method on the RealMCVSR testset. Following \cite{lee2022reference}, we evaluate methods on the task of the first training stage, i.e., using $4\times $ down-sampled ultra-wide and wide-angle videos as the LR and Ref inputs, respectively. We evaluated the models trained in the first stage. 

Table~\ref{table:quan} shows the results. We select several SR methods, i.e., RCAN~\cite{zhang2018rcan}, IconVSR~\cite{chan2021basicvsr}, TTSR\cite{yang2020learning}, DCSR~\cite{wang2021dual}, BasicVSR~\cite{chan2022basicvsr++}, ERVSR~\cite{kim2023efficient} and RefVSR~\cite{lee2022reference}. RCAN, IconVSR and BasicVSR do not utilize a reference, while others are reference-based methods, which is indicated by `R-' in the type column. Only RefVSR and ours in the reference-based methods are video SR methods. {\color{black} We use two variants of our method with different channel numbers for comparison. The smaller one (with `-$small$') is 32, and the other is 64. } The methods with `-pix' in the method column of Table \ref{table:quan} indicate that they are trained with pixel-based loss alone (i.e., the first term in (\ref{eqn:loss1st}) and (\ref{eqn:loss2nd}).) 
 
As with other image restoration/synthesis tasks, SR is affected by the perception-distortion trade-off \cite{blau2018perception}. Specifically, the models trained with pixel-based loss alone tend to yield better quantitative performance, whereas those trained with additional conceptual loss tend to yield better visual quality. For a fair comparison, we show two results for each of the compared methods. To be specific, RCAN, TTSR, DCSR, ERVSR, BasicVSR++ and RefVSR employ $\ell_1$ loss, whereas IconVSR and ours use the Charbonnier loss \cite{charbonnier1994two}, for the pixel-based loss. We can see that our method outperforms all the previous methods in each category, even with fewer parameters (i.e., those with `-$small$'). Both types of models are trained and evaluated with $4\times$ downsampled ultra-wide and wide-angle frames. 

As explained above, we use the wide-angle video as Ref inputs and ultra-wide videos as LR inputs. The wide-angle video frame shares only 50\% of its FoV with the ultra-wide video frame. Previous studies of reference-based SR found that using Ref frames contributes to improving SR quality inside and outside the overlapped FoV. We compute PSNR and SSIM over the pixels belonging to different FoVs to analyze the dependency in image positions, following \cite{lee2022reference}. Table \ref{table:fov} shows the results; 0\%-50\% indicates the overlapped FoV, and 50\%-$r\%$ indicates the centered rectangular FoV having $r\%$ area of the frame minus the overlapped (0\%-50\%) FoV. Like other reference-based methods, our methods also suffer from a performance drop in the non-overlapped FoV. Nevertheless, they still yield better performance than any other method.  The drop ratio in PSNR when comparing the FoV ranges from 0\%-50\% to 50\%-100\% for DCSR and RefVSR is observed to be $-8.5\%$/$-4.2\%$ and $-4.2\%$/$-1.6\%$, respectively. However, for RefVSR++ and RefVSR++-$small$, the reduction is noted to be $-4.1\%$/$-1.4\%$ and $-3.0 \%$/$-1.3\%$. It is notably that RefVSR++ outperforms RefVSR outside the overlapped FoV, which demonstrate the proposed method better utilize temporal Ref information from other frames. 

{\color{black}
To demonstrate the generality of the proposed methods, we evaluate their performance using the Inter4K dataset~\cite{stergiou2021adapool}. Inter4K dataset is composed of 4K resolution videos sourced from the Internet, from which we choose 28 life-themed video clips for our evaluation. To generate narrow-view Ref video inputs, we conduct a center cropping of these videos to a resolution of 1920x1080. Meanwhile, the original videos are resized to 1920x1080 to act as LR inputs. The performance of Ref-based VSR models is presented in Table~\ref{table:inter4k}, our proposed method significantly outperforms others by more than 1.5dB in PSNR. The superior results obtained on the Inter4K dataset demonstrate the adaptability and effectiveness of our proposed method.}

\subsection{Qualitative Evaluation}\label{sec:quality}
\begin{figure*}[!ht]  
\captionsetup[subfigure]{labelformat=empty}

    \centering 
    \begin{tabular}[t]{cccccc}
        \multirow{2}{*}[48pt]{
            \begin{subfigure}{0.14\linewidth}
                \begin{center}
                \includegraphics[width=0.99\linewidth]{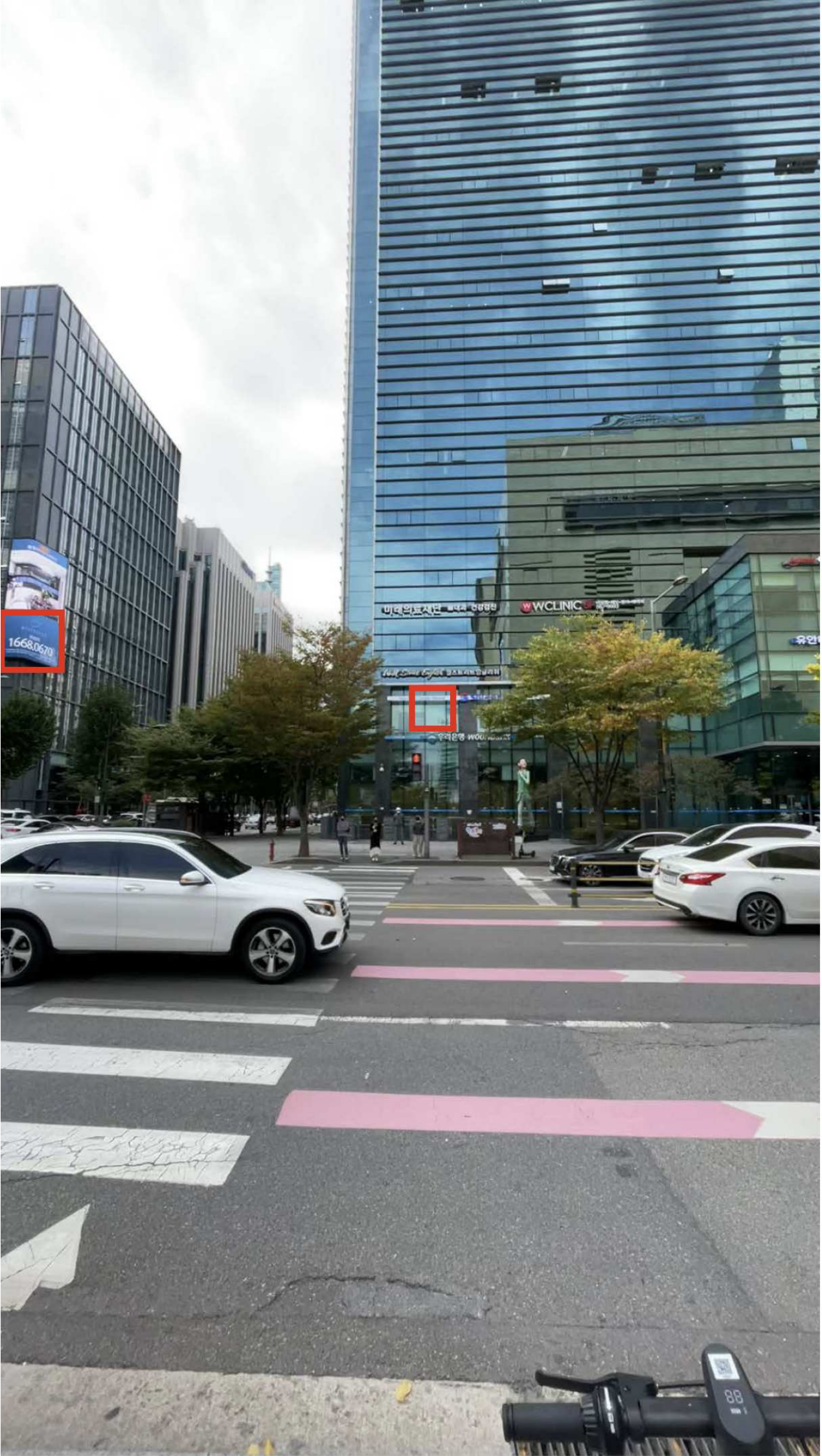}
                \end{center}
            \end{subfigure}
        } &
        \multirow{2}{*}[48pt]{
            \begin{subfigure}{0.14\linewidth}
                \begin{center}
                    \includegraphics[width=0.99\linewidth]{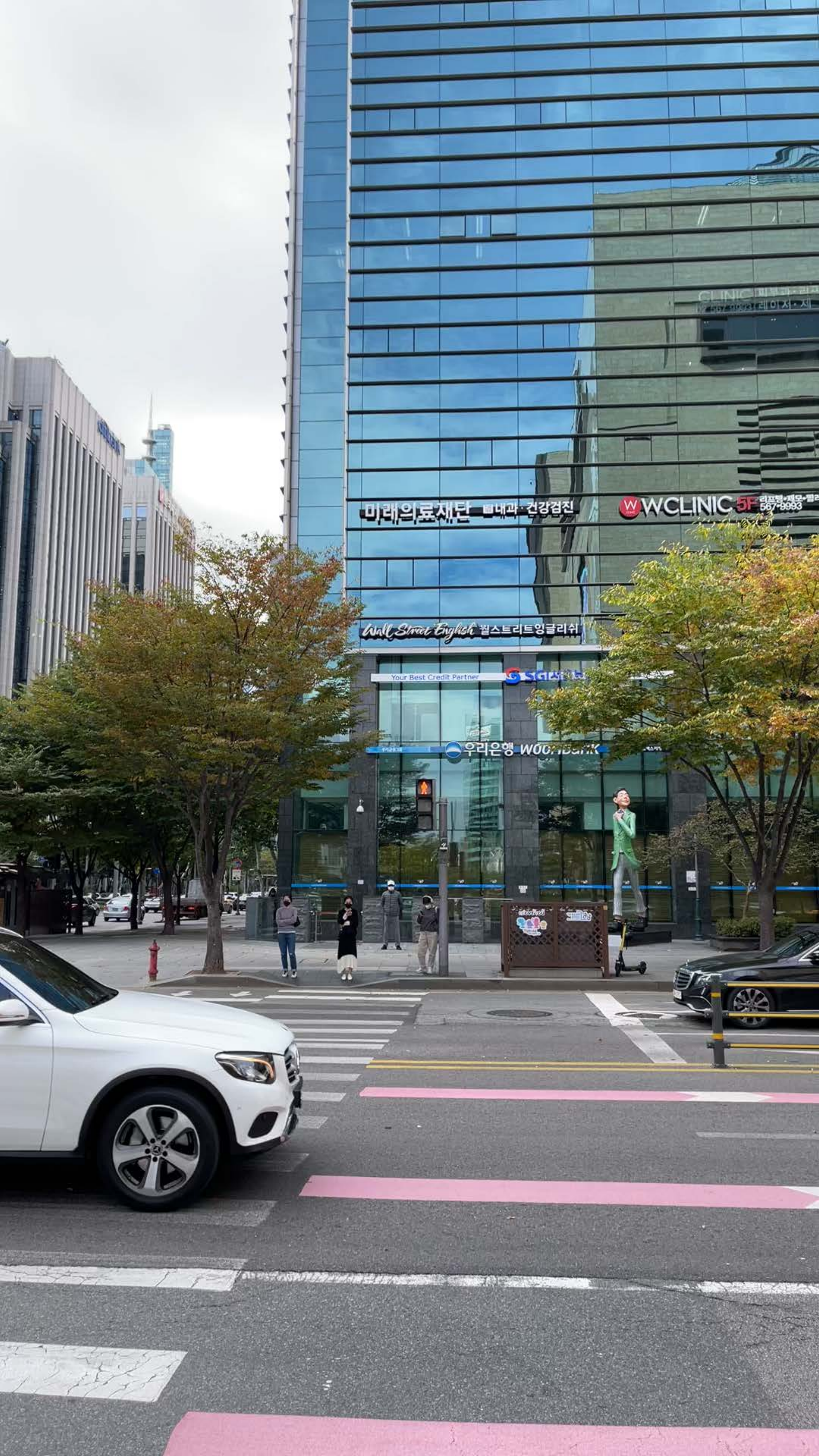}
                \end{center}
            \end{subfigure}
        } &
        \begin{subfigure}{0.14\linewidth}
            \begin{center}
                \includegraphics[width=0.99\linewidth]{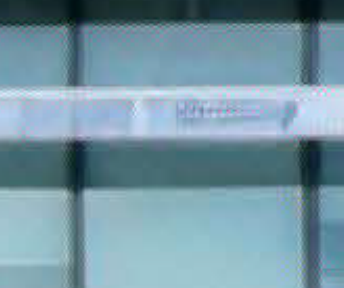}
            \end{center}
        \end{subfigure}
        &
        \begin{subfigure}{0.14\linewidth}
            \begin{center}
                \includegraphics[width=0.99\linewidth]{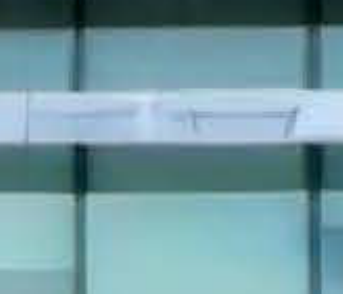}
            \end{center}
        \end{subfigure}
        &
        \begin{subfigure}{0.14\linewidth}
            \begin{center}
                \includegraphics[width=0.99\linewidth]{figures/66_refvsr_01.pdf}
            \end{center}
        \end{subfigure}
        &
        \begin{subfigure}{0.14\linewidth}
            \begin{center}
                \includegraphics[width=0.99\linewidth]{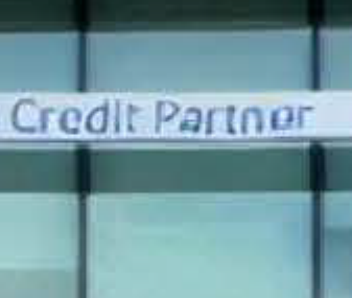}
            \end{center}
        \end{subfigure} \\

    &&        \begin{subfigure}{0.14\linewidth}
            \begin{center}
                \includegraphics[width=0.99\linewidth]{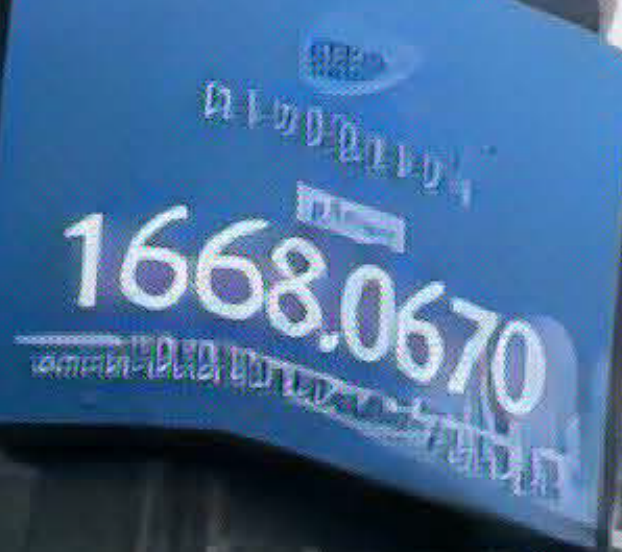}
            \end{center}
        \end{subfigure}
        &
        \begin{subfigure}{0.14\linewidth}
            \begin{center}
                \includegraphics[width=0.99\linewidth]{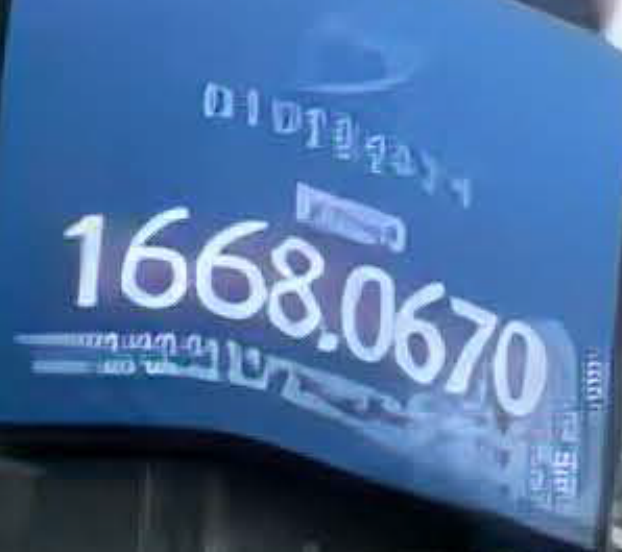}
            \end{center}
        \end{subfigure}
        &
        \begin{subfigure}{0.14\linewidth}
            \begin{center}
                \includegraphics[width=0.99\linewidth]{figures/66_refvsr_02.pdf}
            \end{center}
        \end{subfigure}
        &
        \begin{subfigure}{0.14\linewidth}
            \begin{center}
                \includegraphics[width=0.99\linewidth]{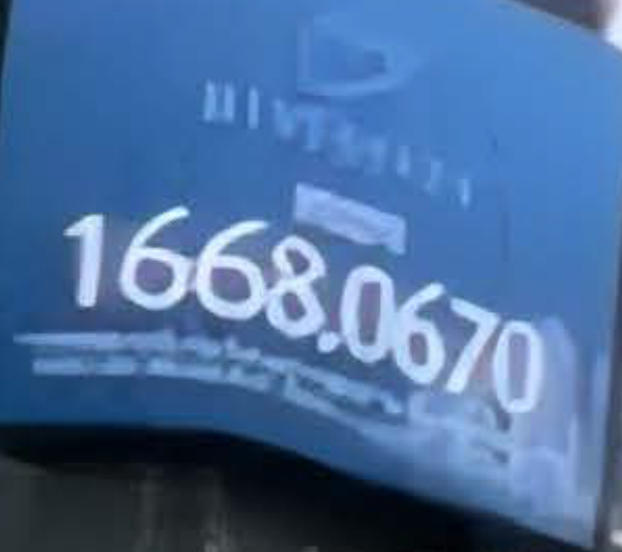}
            \end{center}
        \end{subfigure} \\
        \multirow{2}{*}[48pt]{
            \begin{subfigure}{0.14\linewidth}
                \begin{center}
                    \includegraphics[width=0.99\linewidth]{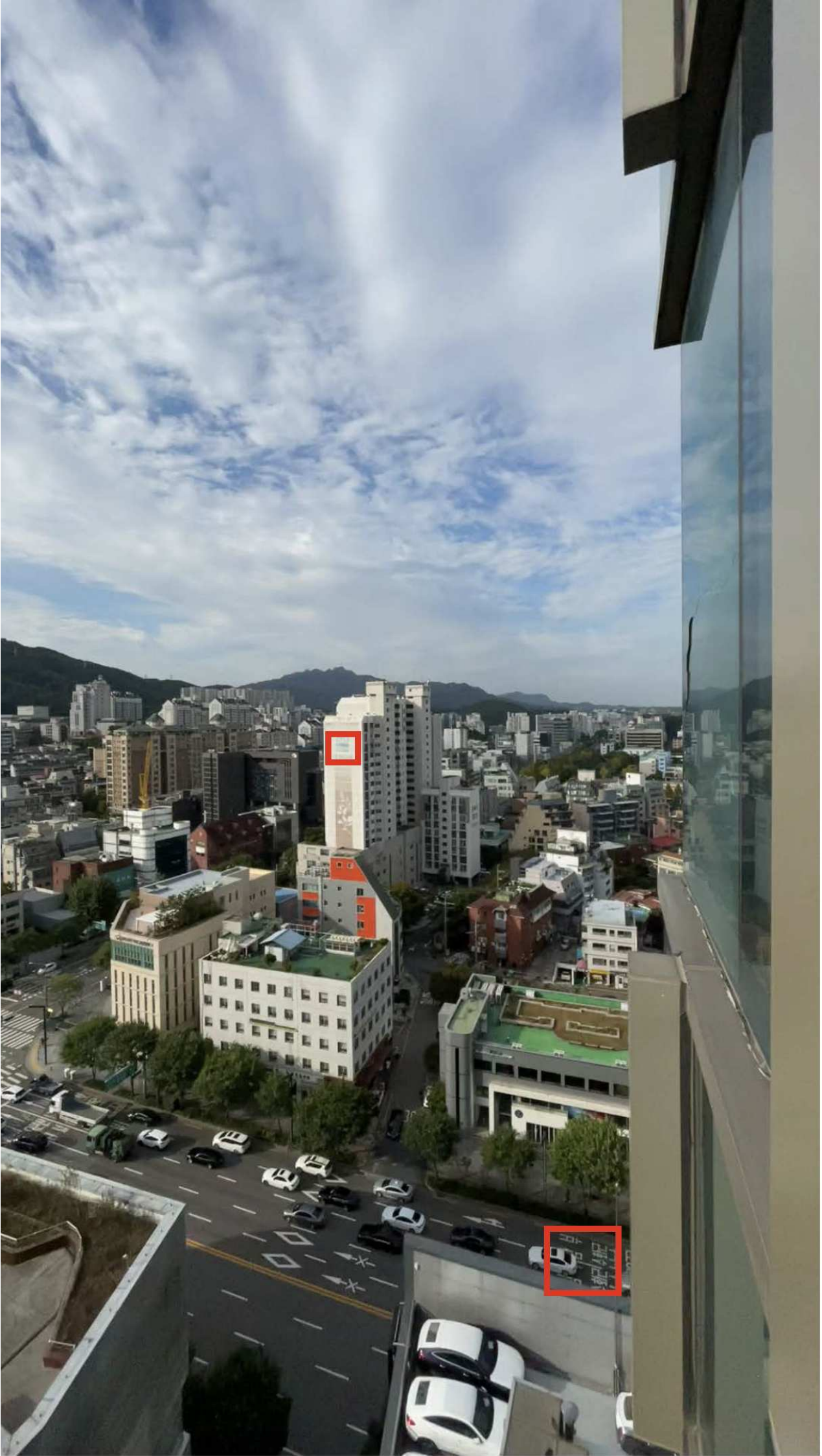}
                    \caption{HR input}
                \end{center}
            \end{subfigure}
        } &
        \multirow{2}{*}[48pt]{
            \begin{subfigure}{0.14\linewidth}
                \begin{center}
                    \includegraphics[width=0.99\linewidth]{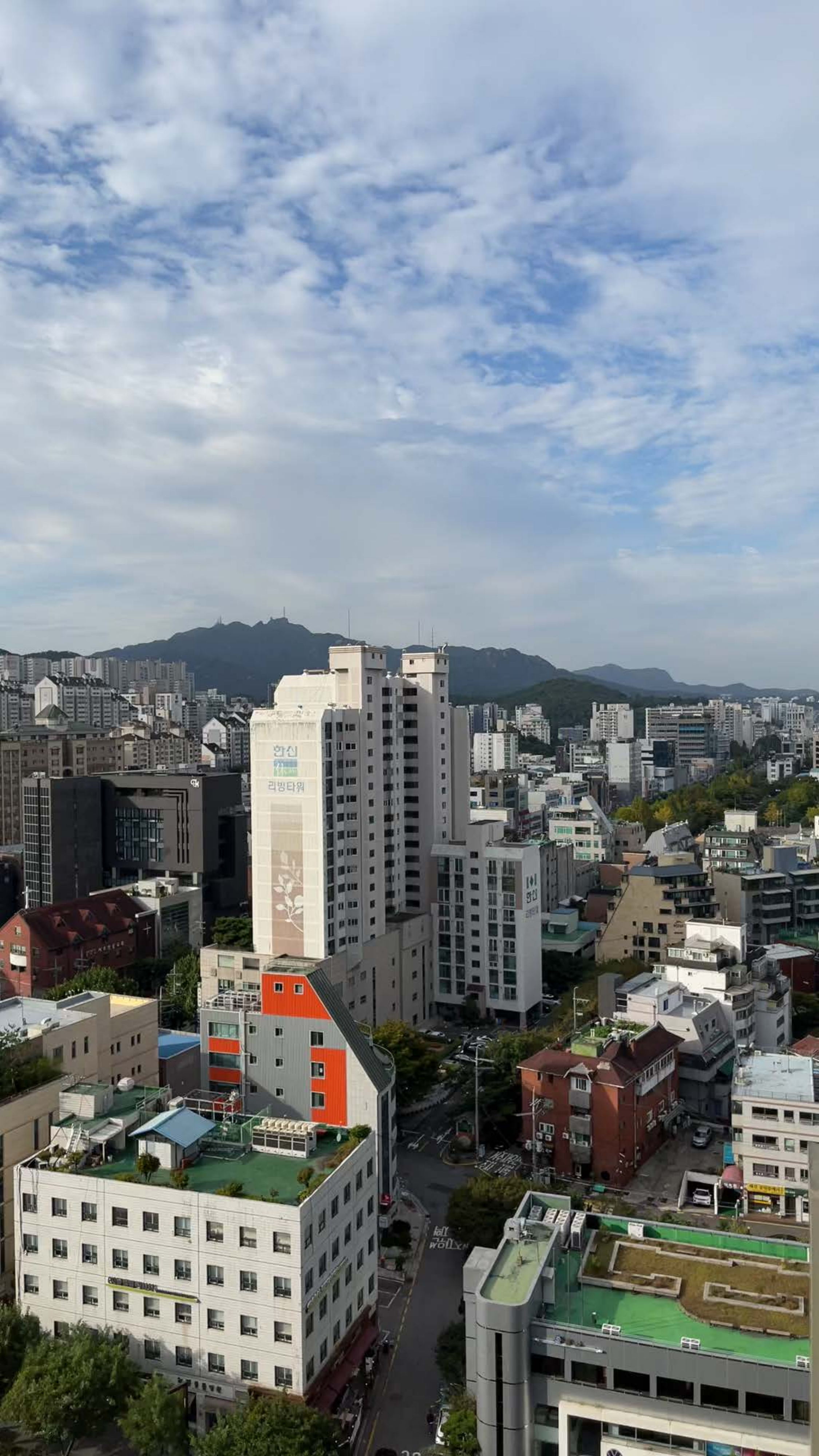}
                    \caption{Ref input}
                \end{center}
            \end{subfigure}
        } &
        \begin{subfigure}{0.14\linewidth}
            \begin{center}
                \includegraphics[width=0.99\linewidth]{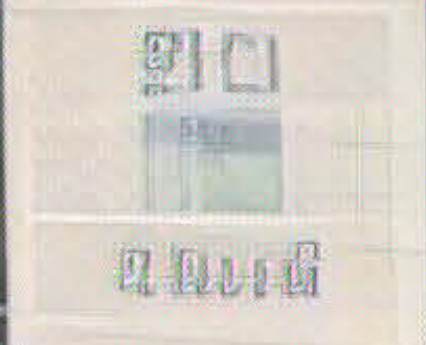}
            \end{center}
        \end{subfigure}
        &
        \begin{subfigure}{0.14\linewidth}
            \begin{center}
                \includegraphics[width=0.99\linewidth]{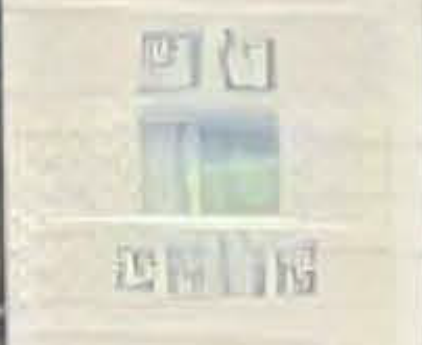}
            \end{center}
        \end{subfigure}
        &
        \begin{subfigure}{0.14\linewidth}
            \begin{center}
                \includegraphics[width=0.99\linewidth]{figures/121_refvsr_01.pdf}
            \end{center}
        \end{subfigure}
        &
        \begin{subfigure}{0.14\linewidth}
            \begin{center}
                \includegraphics[width=0.99\linewidth]{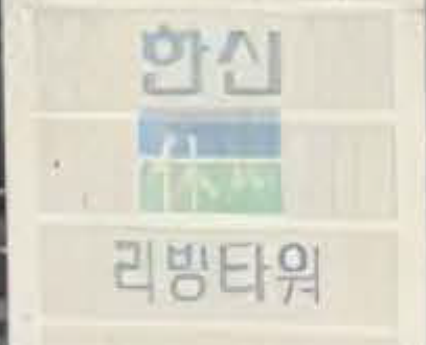}
            \end{center}
        \end{subfigure} \\

    &&        \begin{subfigure}{0.14\linewidth}
            \begin{center}
                \includegraphics[width=0.99\linewidth]{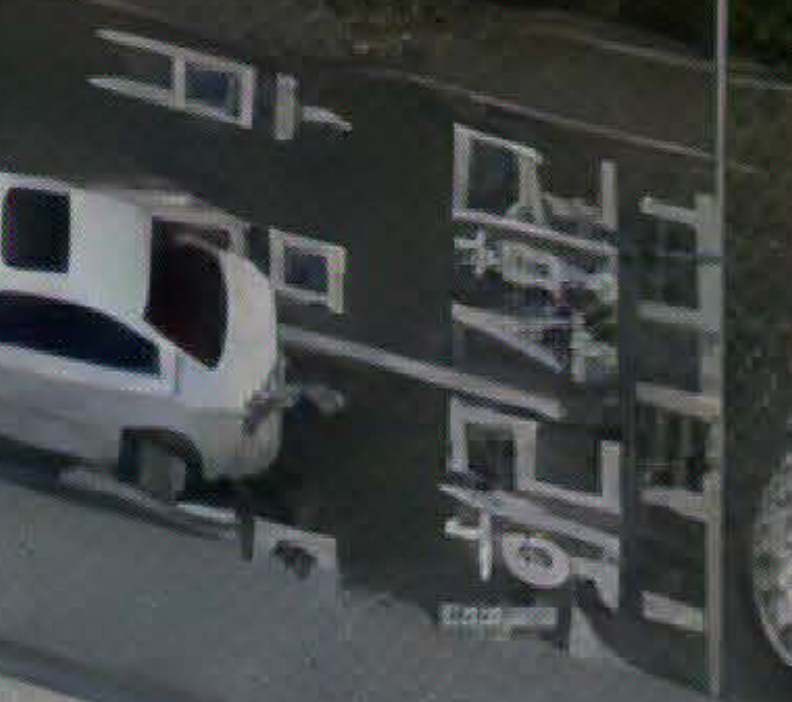}
                \caption{RCAN}
            \end{center}
        \end{subfigure}
        &
        \begin{subfigure}{0.14\linewidth}
            \begin{center}
                \includegraphics[width=0.99\linewidth]{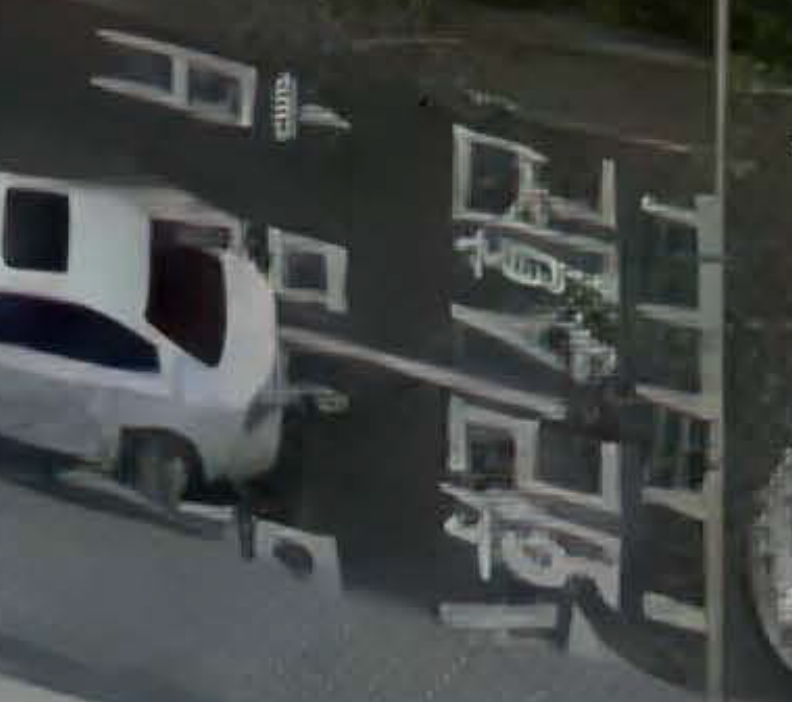}
                \caption{BasicVSR++}
            \end{center}
        \end{subfigure}
        &
        \begin{subfigure}{0.14\linewidth}
            \begin{center}
                \includegraphics[width=0.99\linewidth]{figures/121_refvsr_02.pdf}
                \caption{RefVSR}
            \end{center}
        \end{subfigure}
        &
        \begin{subfigure}{0.14\linewidth}
            \begin{center}
                \includegraphics[width=0.99\linewidth]{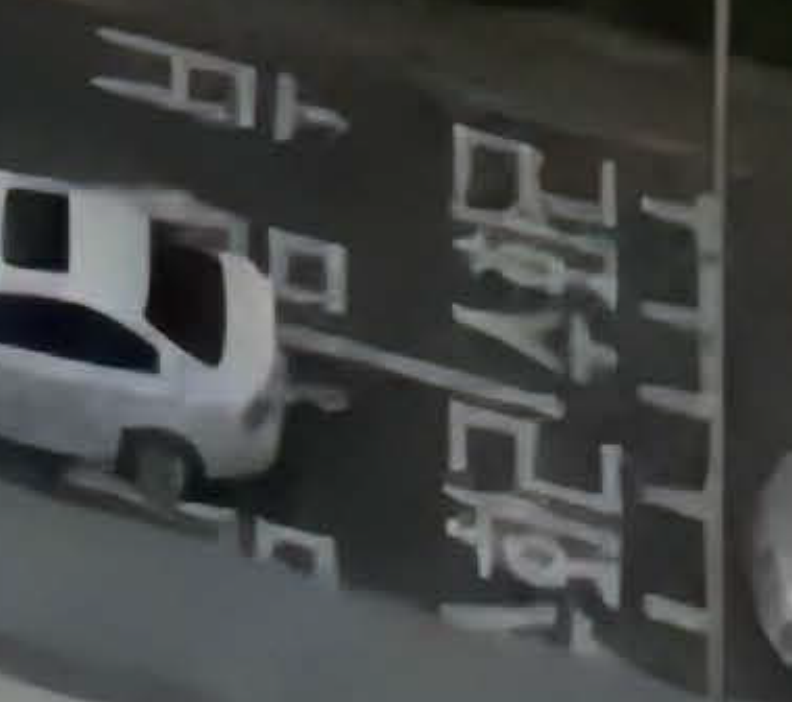}
                \caption{Ours}
            \end{center}
        \end{subfigure} \\
    \end{tabular}
     \vspace{-3mm}
     \caption{\color{black}{Quantitative comparison of different methods on 8K super-resolution. Ground truth image is unavailable.}}
    \label{fig:8k_qua}
\end{figure*}

We use generated 8K videos for qualitative comparison. Figure \ref{fig:8k_qua} shows selected examples of SR outputs generated by different methods. These are the results of $4\times$ super-resolution from the ultra-wide video inputs in the dataset's training split. We select one method from each of the different categories in addition to ours, i.e., RCAN~\cite{zhang2018rcan} (single-image reference SR), BasicVSR++\cite{chan2022basicvsr++} (video SR), and RefVSR\cite{lee2022reference} and ours (reference video SR).  For RCAN and BasicVSR++, we train each using the same training setting as ours on the RealMSVSR dataset. For RefVSR, we used the pre-trained model released by the authors\footnote{\color{red} \url{https://github.com/codeslake/RefVSR}}. We can see that ours generates the clearest textures in the overlapped FoV, including correctly reconstructed numbers and alphabets. In image regions outside the overlapped FoV, it yields the most smooth textures with fewer unnatural artifacts.

\begin{figure}[t]
    \centering
     \begin{minipage}{0.99\columnwidth}
     \begin{subfigure}[b]{0.235\textwidth}
         \centering
         \includegraphics[width=\textwidth]{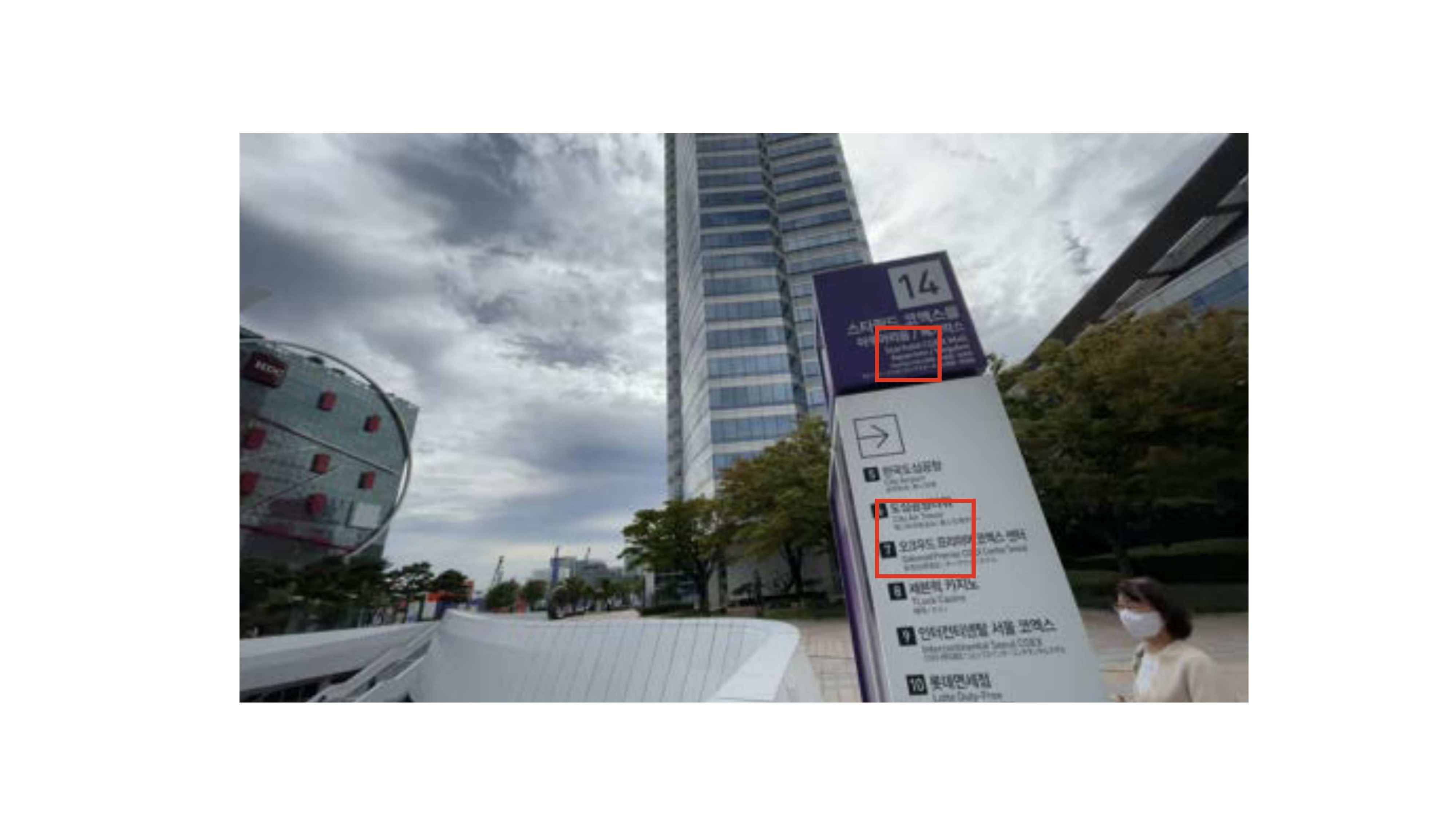}
     \end{subfigure}
     \hfill
     \vspace{0.3mm}
     \begin{subfigure}[b]{0.17\textwidth}
         \centering
         \includegraphics[width=\textwidth]{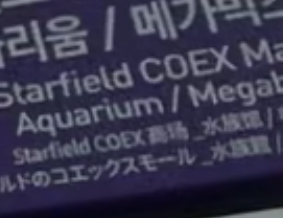}
     \end{subfigure}
     \hfill
     \begin{subfigure}[b]{0.17\textwidth}
         \centering
         \includegraphics[width=\textwidth]{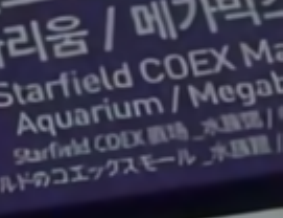}
     \end{subfigure}
     \hfill
     \begin{subfigure}[b]{0.17\textwidth}
         \centering
         \includegraphics[width=\textwidth]{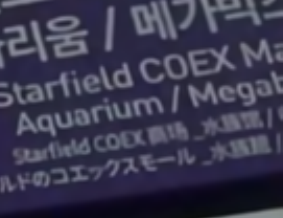}
     \end{subfigure}
     \hfill
     \begin{subfigure}[b]{0.17\textwidth}
         \centering
         \includegraphics[width=\textwidth]{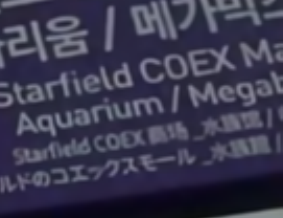}
     \end{subfigure}
     \begin{subfigure}[b]{0.235\textwidth}
         \centering
         \includegraphics[width=\textwidth]{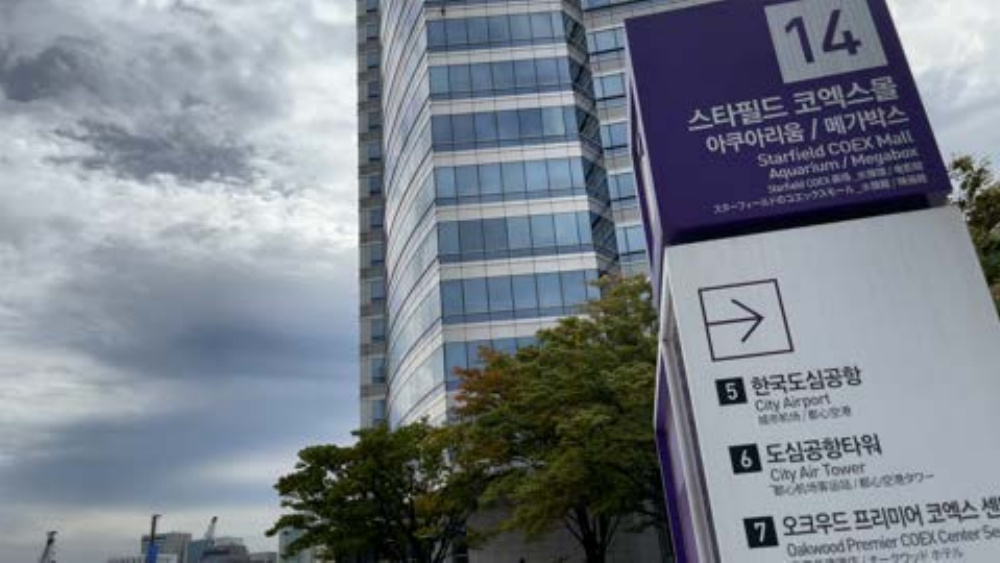}
     \end{subfigure}
     \hfill
     \vspace{0.6mm}
     \begin{subfigure}[b]{0.17\textwidth}
         \centering
         \includegraphics[width=\textwidth]{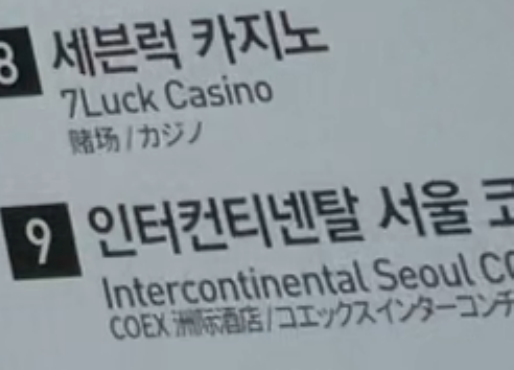}
     \end{subfigure}
     \hfill
     \begin{subfigure}[b]{0.17\textwidth}
         \centering
         \includegraphics[width=\textwidth]{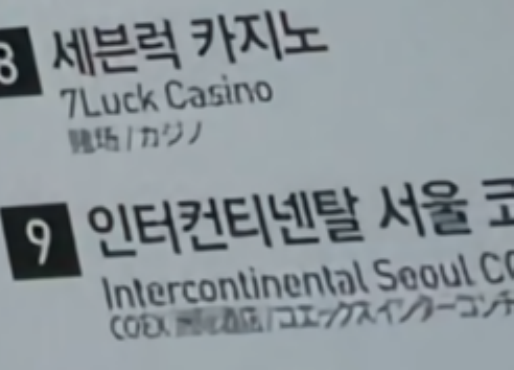}
     \end{subfigure}
     \hfill
     \begin{subfigure}[b]{0.17\textwidth}
         \centering
         \includegraphics[width=\textwidth]{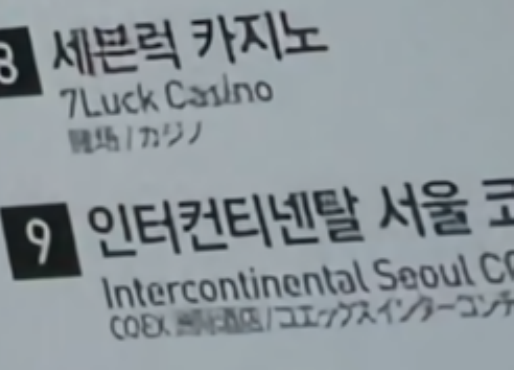}
     \end{subfigure}
     \hfill
     \begin{subfigure}[b]{0.17\textwidth}
         \centering
         \includegraphics[width=\textwidth]{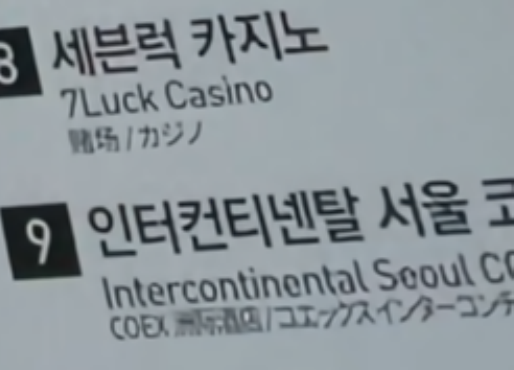}
     \end{subfigure}
     \hfill
     \begin{subfigure}[b]{0.235\textwidth}
         \centering
         \includegraphics[width=\textwidth]{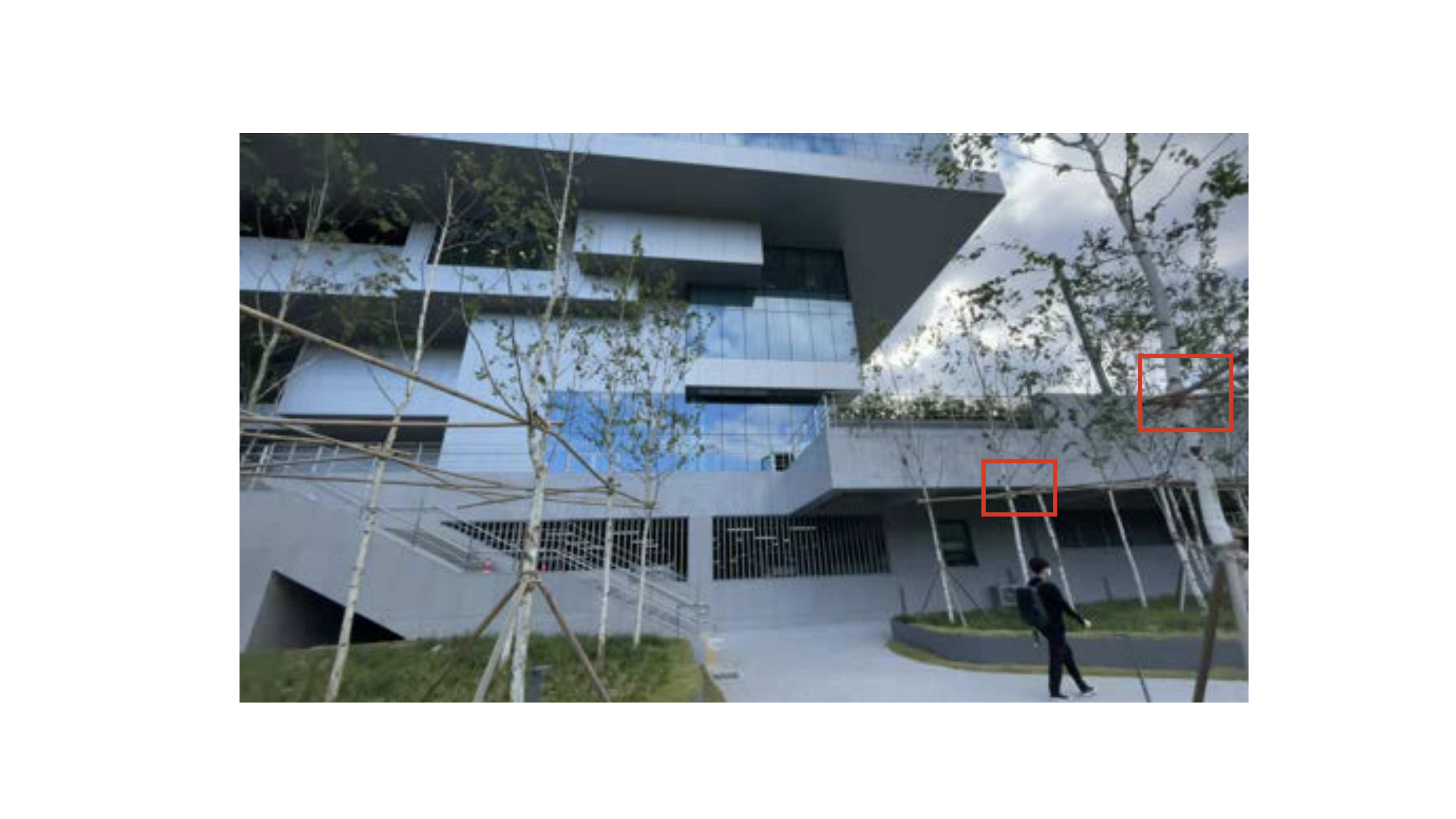}
     \end{subfigure}
     \hfill
     \vspace{0.3mm}
     \begin{subfigure}[b]{0.17\textwidth}
         \centering
         \includegraphics[width=\textwidth]{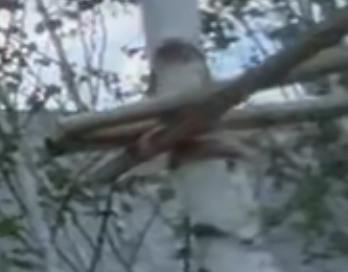}
     \end{subfigure}
     \hfill
     \begin{subfigure}[b]{0.17\textwidth}
         \centering
         \includegraphics[width=\textwidth]{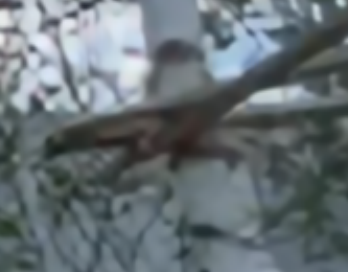}
     \end{subfigure}
     \hfill
     \begin{subfigure}[b]{0.17\textwidth}
         \centering
         \includegraphics[width=\textwidth]{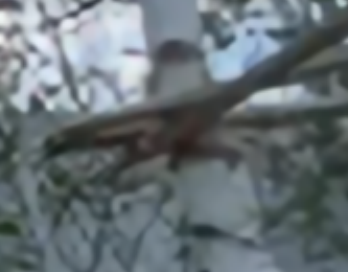}
     \end{subfigure}
     \hfill
     \begin{subfigure}[b]{0.17\textwidth}
         \centering
         \includegraphics[width=\textwidth]{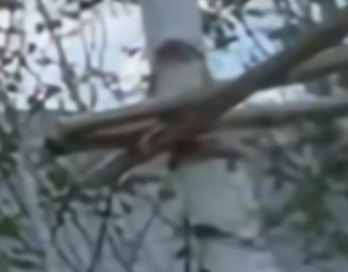}
     \end{subfigure}
     \hfill
     \begin{subfigure}[b]{0.235\textwidth}
         \centering
         \includegraphics[width=\textwidth]{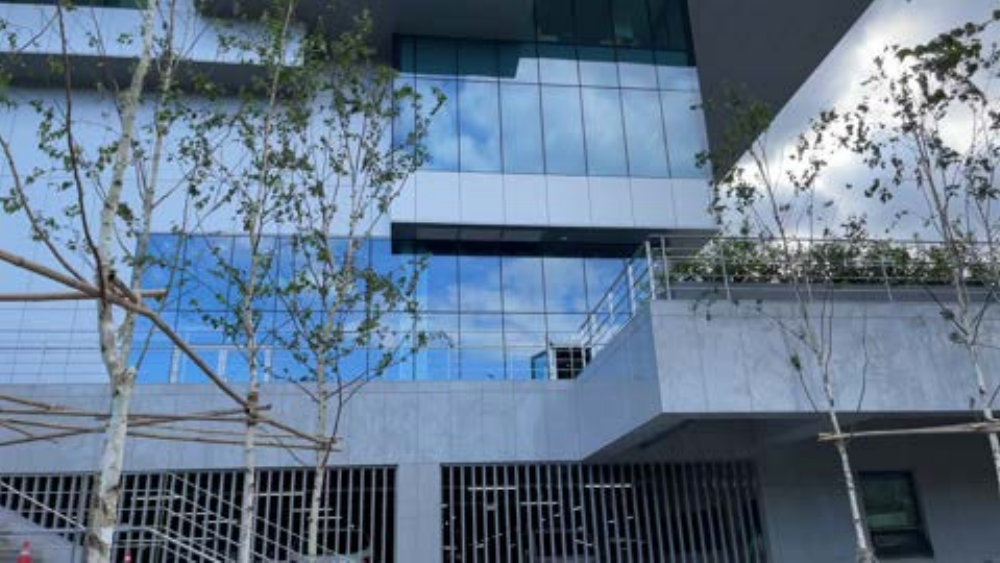}
         \caption{Inputs }
     \end{subfigure}
     \hspace{0.1mm}
     \hfill
     \begin{subfigure}[b]{0.17\textwidth}
         \centering
         \includegraphics[width=0.92\textwidth]{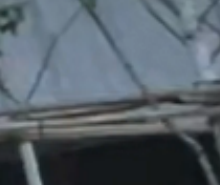}
         \caption{GT }
     \end{subfigure}
     \hfill
     \begin{subfigure}[b]{0.17\textwidth}
         \centering
         \includegraphics[width=0.92\textwidth]{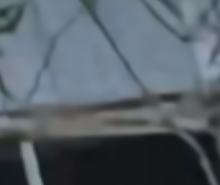}
         \caption{Baseline }
     \end{subfigure}
     \hfill
     \begin{subfigure}[b]{0.17\textwidth}
         \centering
         \includegraphics[width=0.92\textwidth]{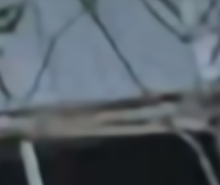}
         \caption{SR }
     \end{subfigure}
     \hfill
     \begin{subfigure}[b]{0.17\textwidth}
         \centering
         \includegraphics[width=0.92\textwidth]{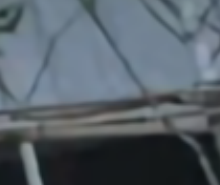}
         \caption{SR +Ref }
     \end{subfigure}
        \caption{Qualitative comparison of the model variants shown in Table \ref{table:ablation_dual}. Left most two columns show input ultra-wide and wide-angle images, respectively.}
     \label{fig:qua_comp}
     \end{minipage}
\end{figure}

\subsection{Ablation Study}
\label{sec:ablation}

We conduct an ablation test to identify the effectiveness of each component of our method. We experimentally evaluate several variants of the proposed network with different ablated components. Here, we use Charbonnier loss for simplicity. The results are shown in Table~\ref{table:ablation_dual}. 

The variant in the first row of Table~\ref{table:ablation_dual} does not use Ref frames to improve the SR quality. The variant in the second row serves as the baseline model, which primarily aggregates basic features and relies only on optical-flow-based alignment for aligning adjacent frames (i.e., $t-1$ and $t$). Its performance is 35.19dB.
The third-row variant incorporates a confidence map and mimics the feature propagation mechanism of RefVSR \cite{lee2022reference}, resulting in relatively lower performance. The fourth-row variant, which integrates the proposed SR stream, shows an improvement in PSNR by 0.53dB. This highlights the effectiveness of the DCN-based feature alignment.
The fifth and sixth variants build upon the baseline by adding the proposed Ref feature stream. The sixth variant, in particular, further integrates Ref residual propagation. The sixth variant's performance, surpassing that of the fifth variant, underscores the efficacy of residual propagation discussed in Sec.~\ref{sec:residual}. 
It improves upon the baseline's PSNR by 0.43dB, affirming the effectiveness of the Ref  stream.
The variant in the last row represents our full network, which integrates both the proposed SR stream and Ref stream. It achieves the highest PSNR of 35.90dB. Figure \ref{fig:qua_comp} shows qualitative comparisons of proposed components, which align with these quantitative comparisons.

\section{Computational Cost and Limitations}

Table~\ref{table:cost} presents the computational costs of different Ref-based VSR methods in the setting where we input frames with the size of $480\times 270$ pixels.  RefVSR exhibits the longest runtime and highest GMACs requirement due to its multi-to-one architecture.  The proposed RefVSR++ and RefVSR++-$small$ demonstrate significant improvements in PSNR with only a modest increase in runtime compared to ERVSR. However, both RefVSR and our method 
 need more memory,
which is
a limitation of such a design.

\section{Summary and Conclusion}

We have proposed a new method for reference-based video super-resolution (SR) aiming at super-resolving videos captured by a smartphone's multi-camera system.  The problem is formulated as super-resolving an input low-resolution (LR) video with the help of an input reference (Ref) video.  We assume that the Ref video has a narrower FoV and thus contains higher-resolution images of a part of the scene. The proposed method, RefVSR++,  extends RefVSR, the existing method for the problem, in two aspects. One is to add a parallel stream for aggregating Ref image features over time in addition to the standard stream aggregating the LR and Ref features over time. This mechanism can derive richer information from the two inputs. The outputs of the two streams are integrated to produce the super-resolution image of the LR input at each time step. The other is an improved mechanism for aligning the LR and Ref features across the neighboring time steps. We showed through experiments the effectiveness of RefVSR++. 

\section*{Acknowledgments}
\noindent
This work was partly supported by JSPS KAKENHI Grant Number 20H05952 and 23H00482.

{\small
\bibliographystyle{ieee_fullname}
\bibliography{egbib}
}

\end{document}